\documentclass[runningheads]{llncs}

\usepackage{eccv}

\usepackage{eccvabbrv}

\usepackage{graphicx}
\usepackage{booktabs}

\usepackage{titletoc} 
\usepackage{float}
\usepackage{graphicx}
\usepackage{multirow}
\usepackage{amsmath}
\usepackage{xcolor} 
\usepackage{caption}
\usepackage{algorithm}
\usepackage{algorithmic}
\usepackage{pifont} 
\usepackage{color, colortbl} 
\usepackage{arydshln} 
\usepackage{tabularx} 
\usepackage{footnote} 
\usepackage{booktabs} 

\usepackage[accsupp]{axessibility}  

\usepackage{hyperref}

\usepackage{orcidlink}

\newcommand\blfootnote[1]{%
  \begingroup
  \renewcommand\thefootnote{}%
  \footnotetext{#1}%
  \endgroup
}

\newcommand{\appref}[1]{\hyperref[#1]{Appendix~\ref*{#1}}}

\begin{document}

\title{GenRecal: Generation after Recalibration\\from Large to Small Vision-Language Models} 

\titlerunning{GenRecal: Generation after Recalibration}

\author{
\renewcommand{\arraystretch}{1.0}\setlength{\tabcolsep}{0pt}
\begin{tabular}{@{}*{3}{>{\centering\arraybackslash}p{0.333\textwidth}}@{}}
Byung-Kwan Lee$^{\dagger}$ & Ryo Hachiuma & Yong Man Ro \\
{\small NVIDIA} & {\small NVIDIA} & {\small KAIST} \\
{\small\texttt{byungkwanl@nvidia.com}} & {\small\texttt{rhachiuma@nvidia.com}} & {\small\texttt{ymro@kaist.ac.kr}} \\[1.3em]
\multicolumn{3}{@{}c@{}}{%
  \begin{tabular}{@{}*{2}{>{\centering\arraybackslash}p{0.42\textwidth}}@{}}
  Yu-Chiang Frank Wang & Yueh-Hua Wu \\
  {\small NVIDIA} & {\small NVIDIA} \\
  {\small\texttt{frankwang@nvidia.com}} & {\small\texttt{krisw@nvidia.com}} \\
  \end{tabular}%
} \\
\end{tabular}
}

\authorrunning{B.-K. Lee, R. Hachiuma, et al.}
\tocauthor{Byung-Kwan Lee, Ryo Hachiuma, Yong Man Ro, Yu-Chiang Frank Wang, Yueh-Hua Wu}

\institute{}

{\maketitle 
\centering
\vspace{-3mm}
\includegraphics[width=\textwidth]{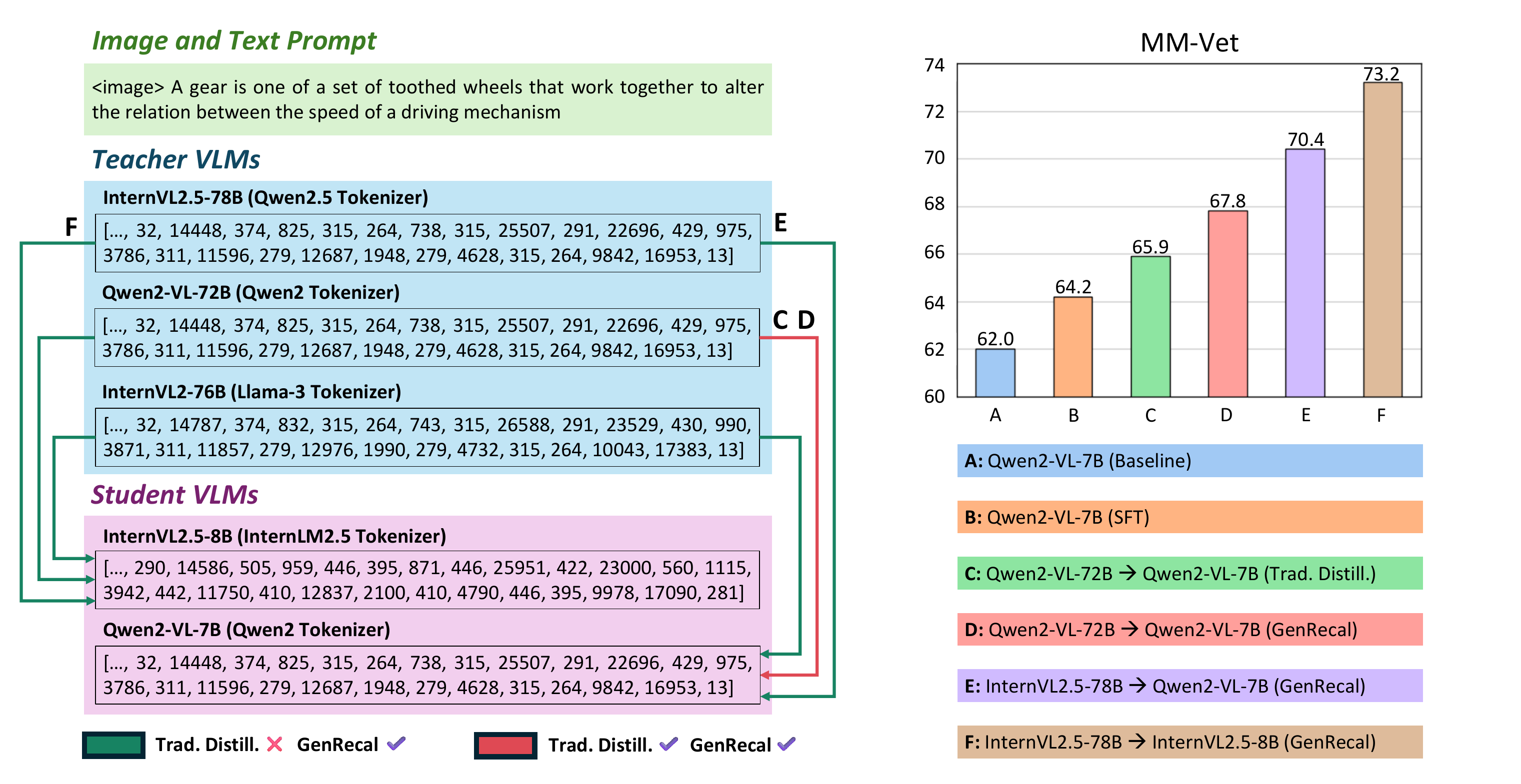}
\vspace{-7mm}
\captionof{figure}{(Left) Visualizing the token indices of a given image and text prompt and representing the possibility of distillation among various VLM pair combinations, comparing traditional distillation with our proposed distillation framework, GenRecal. Note that the parentheses mean each VLM's LLM tokenizer, `...' indicates the placement of image features, and the number of these features varies depending on the image embedding strategy. (Right) Comparing the performance of a challenging evaluation benchmark, MM-Vet~\cite{yu2023mm}, with [A] baseline, [B] SFT on the baseline, [C] traditional distillation and [D] GenRecal from same token types of large VLMs, and GenRecal with more powerful [E] large and [F] small VLMs.
}
\label{fig:1}
\vspace{-3mm}
}
\begin{abstract}
\blfootnote{$^{\dagger}$~Project Lead}%
Recent advancements in vision-language models (VLMs) have leveraged large language models (LLMs) to achieve performance on par with closed-source systems like GPT-4V. However, deploying these models in real-world scenarios, particularly on resource-constrained devices, remains challenging due to their substantial computational demands. This has spurred interest in distilling knowledge from large VLMs into smaller, more efficient counterparts. A key challenge arises here from the diversity of VLM architectures, which are built on different LLMs and employ varying token types—differing in vocabulary size, token splits, and token index ordering. To address this challenge of limitation to a specific VLM type, we present \textbf{Gen}eration after \textbf{Recal}ibration (GenRecal), a general-purpose distillation framework for VLMs. GenRecal incorporates a Recalibrator that aligns and adapts feature representations between heterogeneous VLMs, enabling effective knowledge transfer across different types of VLMs. Through extensive experiments on multiple challenging benchmarks, we demonstrate that GenRecal significantly improves baseline performances, eventually outperforming large-scale open- and closed-source VLMs.
\end{abstract}    
\section{Introduction}

\begin{figure}[t!]
    \vspace{-3mm}
    \centering
    \includegraphics[width=\textwidth]{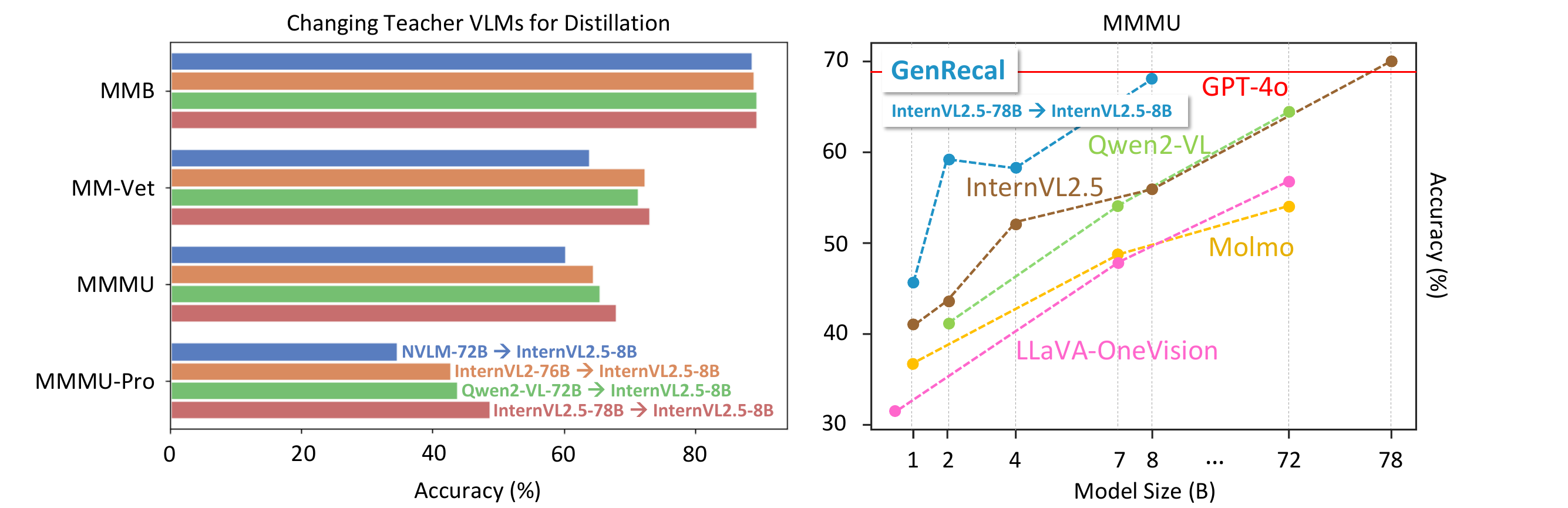}
    \vspace{-7mm}
    \caption{(Left) Comparison of the challenging benchmark performances, MMB~\cite{liu2023mmbench}, MM-Vet~\cite{yu2023mm}, MMMU~\cite{yue2023mmmu}, and MMMU-Pro~\cite{yue2024mmmu} by changing large VLMs. The more powerful large VLMs we select, the greater the performance improvement we can achieve. (Right) Comparing the performance of the challenging benchmark: MMMU~\cite{yue2023mmmu}, with GenRecal and various VLMs across model sizes. Note that all the experiments in \cref{fig:1} and \cref{fig:2} are conducted on the equal training dataset.}
    \label{fig:2}
    \vspace{-5mm}
\end{figure}

\label{sec:intro}
Vision-language models (VLMs) have emerged as powerful tools for understanding and processing multimodal information, enabling tasks such as image captioning and visual question answering~\cite{tan2019lxmert, blip}. Recent advancements in VLMs~\cite{alayrac2022flamingo, blip2, liu2023improved} have leveraged large-scale language models (LLMs) to enhance reasoning and generation capabilities by incorporating extensive textual knowledge~\cite{2019t5, soldaini-etal-2024-dolma}. In pursuit of superior performance, state-of-the-art VLMs~\cite{li2024llava, wang2024qwen2vl, deitke2024molmo, nvlm2024, chen2024expanding} now integrate scaled-up LLMs with up to 72B parameters, achieving results comparable to proprietary models such as GPT-4V~\cite{gptsyscard} and Claude-3.5 Sonnet~\cite{claude3series2024}.

However, the increasing scale of recent VLMs introduces substantial computational overhead, limiting their practicality in real-world scenarios—particularly for on-device deployment. To mitigate this challenge, recent work has explored distillation techniques~\cite{cai2024llava, shu2024llava, feng2024align} that transfer knowledge from large (teacher) VLMs to smaller (student) ones. Yet, existing distillation methods suffer from a fundamental limitation: they typically assume that teacher and student produce sequences of equal token length, enabling token-level distance metrics such as KL divergence. This assumption breaks down when the two models use different tokenizers or adopt different image-splitting mechanisms (see \appref{app:A}), making the distillation process infeasible. Consequently, the range of viable teacher-student VLM pairs becomes severely restricted (see \appref{app:B}). In other words, current distillation approaches only operate when the teacher and student share identical vocabulary sizes, token splits, and token-index-ordering schemes (collectively referred to as token types), as illustrated on the left side of \cref{fig:1}. These differences naturally lead to mismatched input or output token lengths across VLMs, preventing effective distillation. Note that even VLMs within the same family may employ different token types, further hindering compatibility.

\begin{figure}[t!]
    \vspace{-3mm}
    \centering
    \includegraphics[width=\textwidth]{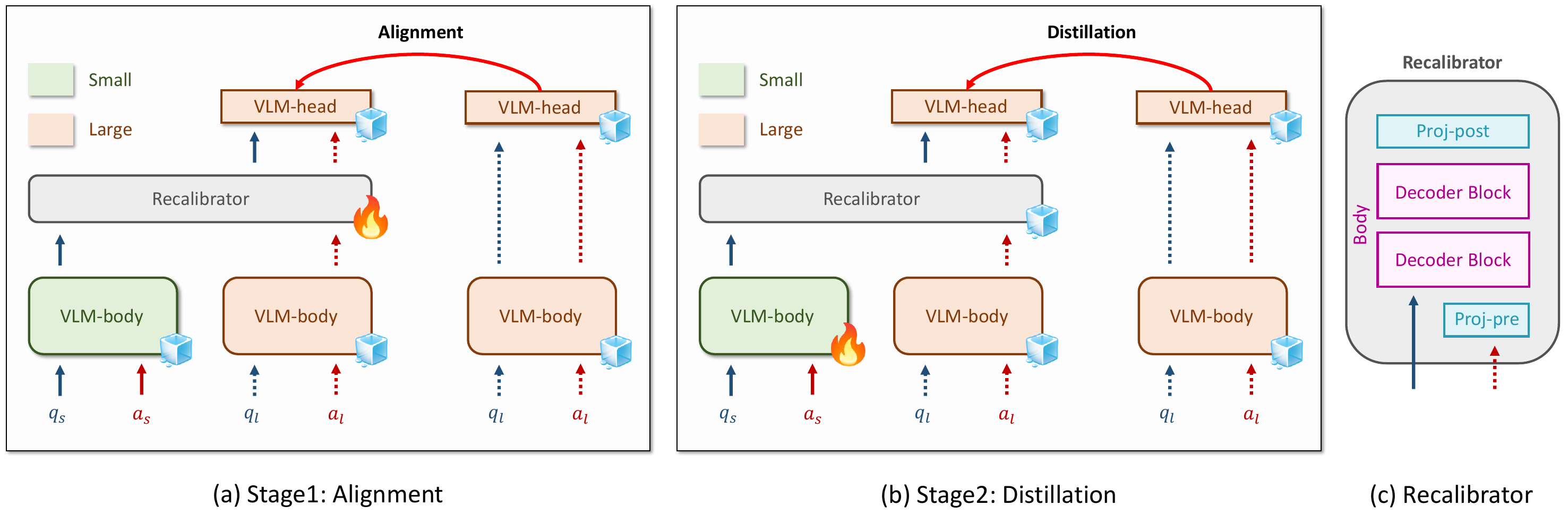}
    \vspace{-7mm}
    \caption{
    Overview of the GenRecal architecture and its training stages. We denote the question and answer tokens from the small VLM as $q_s$ and $a_s$, and those from the large VLM as $q_l$ and $a_l$. For simplicity, the vision encoder and image-token projector are omitted from the illustration. Note that Recalibrator is used only during training as a bridge between heterogeneous VLMs. During inference, Recalibrator is removed, leaving the small VLM’s architecture unchanged and no extra computational cost.
    }
    \label{fig:3}
    \vspace{-5mm}
\end{figure}

To overcome this limitation, we propose \textbf{Gen}eration after \textbf{Recal}ibration (GenRecal), a general-purpose distillation framework for token types-agnostic VLM distillation. At its core, GenRecal introduces a \textit{Recalibrator} that aligns and adapts the feature representations of small VLMs with those of large VLMs. This design is inspired by prior studies demonstrating that word embeddings from different models can be linearly mapped into a shared embedding space~\cite{mikolov2013exploiting, smith2017offline, conneau2017word}. Unlike approaches that align only word embeddings, Recalibrator performs joint visual–linguistic alignment on the hidden representations before the language head. By projecting small VLM features into large VLMs’ feature space as shared representation, Recalibrator effectively bridges the gap between the large and small VLMs, enabling general-purpose knowledge transfer.

To demonstrate the necessity of GenRecal, we first compare its performance with that of traditional distillation, as shown in experiment types C and D of \cref{fig:1}. Although both experiments employ the same token types of VLMs—Qwen2-VL-72B as the teacher and Qwen2-VL-7B as the student—GenRecal consistently outperforms the traditional distillation method implemented by LLaVA-KD~\cite{cai2024llava}. This result highlights that aligning feature representations between large and small VLMs is crucial for minimizing information loss during distillation, even when the models are homogeneous. Moreover, when we replace the large VLM from Qwen2-VL-72B with the more powerful InternVL2.5-78B, GenRecal achieves enhanced performance. Subsequently, replacing the small VLM with InternVL2.5-8B leads to an even stronger model, as summarized on the right side of \cref{fig:1}. These experiments highlight the significance of GenRecal because traditional methods cannot handle VLM pairs with different token types.

Our contributions can be summarized as follows:

\begin{itemize}
    \item \textbf{Token Types-agnostic Recalibration:} GenRecal employs the Recalibrator to align and adapt the feature representations of large and small VLMs, enabling general-purpose distillation across token types that differ in vocabulary size, token splits, and token-index ordering. 
    \item \textbf{Broad Applicability:} GenRecal is compatible with a wide range of VLM architectures across different model sizes, overcoming the challenge of selecting large VLMs that have different token types and demonstrating its practicality for real-world deployment in resource-constrained settings.
\end{itemize}
\section{Related Work}

\begin{figure}[t!]
    \vspace{-3mm}
    \centering
    \includegraphics[width=0.6\textwidth]{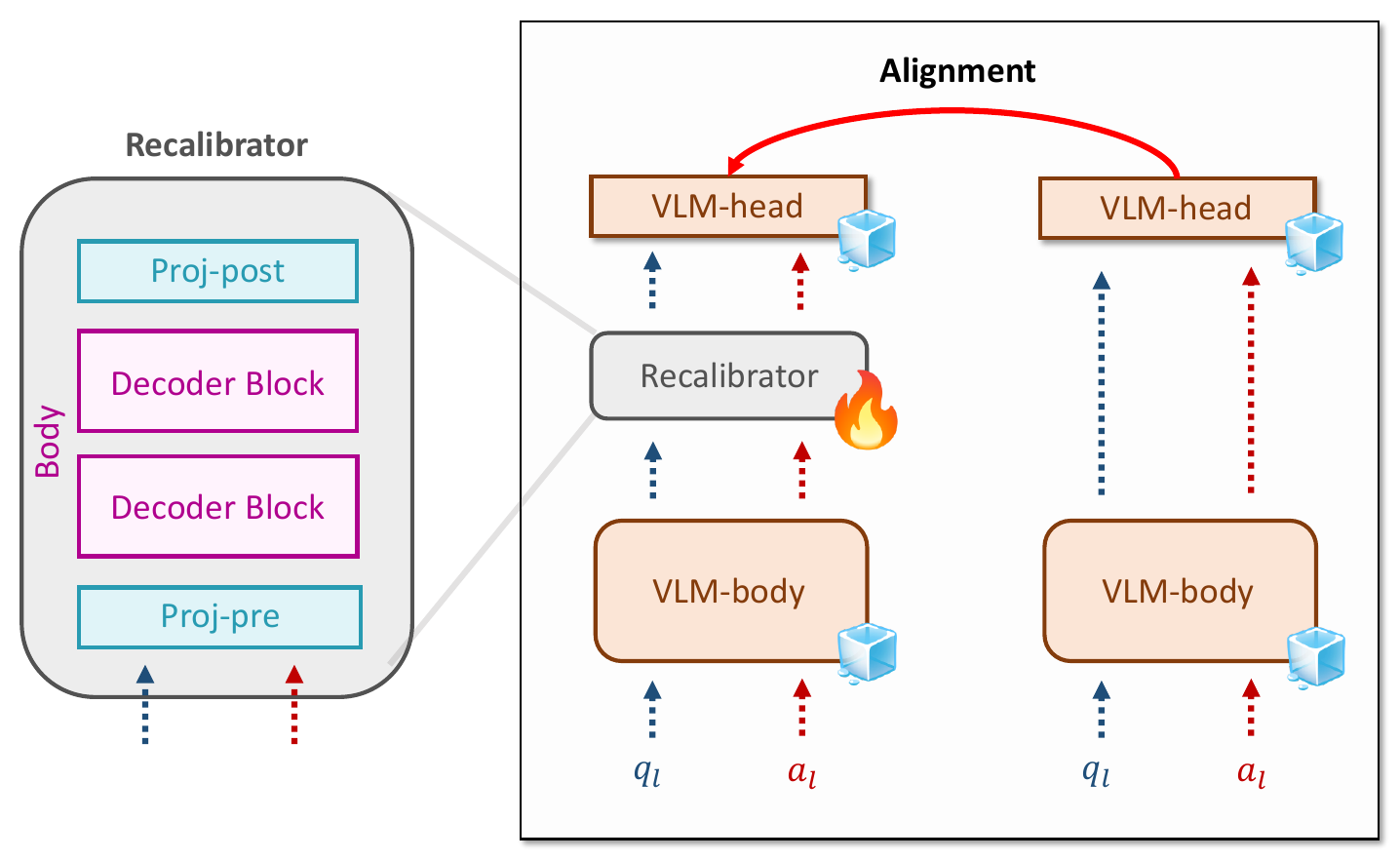}
    \vspace{-3mm}
    \caption{Overview of regularization simultaneously done with first training stage-alignment. Note that its propagation rule is different with \cref{fig:3}(c) where Recalibrator takes small VLM's tokens for the question and large VLM's tokens for the answer. In contrast, in this figure, only large VLM’s tokens (question and answer) are fed into Recalibrator. Basically, \textit{Rec-body} has the hidden dimension of the small VLM, so we should change this hidden dimension of large VLM's tokens to that of small VLM's ones via \textit{Proj-pre} when the large VLM's tokens are fed into Recalibrator.}
    \label{fig:4}
    \vspace{-3mm}
\end{figure}

Large-scale VLMs such as NVLM-72B~\cite{nvlm2024}, Qwen2-VL-72B~\cite{wang2024qwen2vl}, and InternVL2.5-78B~\cite{chen2024expanding} have recently approached the performance of GPT-4V~\cite{gptsyscard} and Claude-3.5 Sonnet~\cite{claude3series2024} (see \appref{app:C} for the evolution of VLMs). However, they impose a significant computational burden in real-world applications, such as on-device processing. Hence, it is necessary to develop small VLMs for deployment on lightweight devices, and this demand has led to two types of distillation approaches. The first constructs visual instruction tuning datasets~\cite{li2024llava, gu2024infinity, zhang2024beyond, li2024cvlm, chen2023can, hu2024mplug, wang2024enhancing, Sujet-Finance-QA-Vision-100k} from large-scale VLMs and trains small VLMs on these curated datasets. The second performs feature distillation: LLaVA-MoD~\cite{shu2024llava} conducts logit-based distillation on a mixture-of-experts (MoE)~\cite{shazeer2017, riquelme2021scaling} architecture. LLaVA-KD~\cite{cai2024llava} also uses logit-based distillation but adopts a three-stage training process in which the trainable parameters differ at each stage to effectively warm up and distill the knowledge. Align-KD~\cite{feng2024align} distills both vision-encoder features and decoder logits, while encouraging the student VLM's first decoder layer's attention map to resemble that of the large VLM. MoVE-KD~\cite{cao2025move} employs multiple vision encoders for the large VLM and applies a mixture of LoRA~\cite{hu2021lora} to the small VLM, distilling visual information through visual attention maps. Besides, there are earlier distillation studies (see \appref{app:C}), but they consider neither token-derived nor different-token-type distillation.

However, existing works are largely constrained to VLMs sharing identical token types, including vocabulary size, token segmentation, and token index ordering. When the large and small VLMs employ different tokenizers or image embedding strategies, distillation becomes problematic—KL divergence cannot be computed due to mismatched output token counts, and token indices may no longer correspond to semantically equivalent meaning. Overcoming this limitation would represent a significant advancement, broadening the applicability and robustness of distillation techniques for diverse VLM architectures.
\section{GenRecal: Generation after Recalibration}
\label{sec:method}

\subsection{Model Architecture and Components}
We briefly illustrate the overall model architecture and training procedure in \cref{fig:3}. GenRecal consists of three main components: a large (teacher) VLM, a small (student) VLM, and a Recalibrator module. For the large VLMs, we select models with over 72B parameters to ensure competitive performance: NVLM-72B (Qwen2-72B)~\cite{nvlm2024}, Qwen2-VL-72B (Qwen2-72B)~\cite{wang2024qwen2vl}, InternVL2-76B (Llama3-70B)~\cite{chen2024far}, and InternVL2.5-78B (Qwen2.5-72B)~\cite{chen2024expanding}. For the small VLMs, we adopt various model sizes, including Qwen2-VL-2B/7B (Qwen2-2B/7B) and InternVL2.5-1B/2B/4B/8B (InternLM2.5-1.8B/7B and Qwen2.5-0.5B/3B/72B). Note that the parentheses denote the LLMs used within each VLM. Throughout this section, we divide a VLM architecture into four key modules: the vision encoder, vision projector, \textit{VLM-body}, and \textit{VLM-head} (\ie language head). Recalibrator is designed to acquire shared feature representations within a common latent space. It consists of two decoder blocks and two projection layers, as depicted in \cref{fig:3}(c). We denote the decoder blocks as \textit{Rec-body}, and the projection layers as \textit{Proj-pre} and \textit{Proj-post}. The decoder blocks, \textit{Rec-body}, adopt the same structural design as the student VLM’s decoder, while both \textit{Proj-pre} and \textit{Proj-post} are implemented as single linear layers.

{
\setlength{\textfloatsep}{0pt}
\begin{algorithm}[t!]
\caption{First stage Loss Functions (Alignment)}
\begin{algorithmic}[1]
    \STATE \textbf{Input:} ($q_l$, $a_l$), ($q_s$, $a_s$), $gt_l$ (label index)
    \STATE $[z_{q_l}, z_{a_l}]\gets$\textit{VLM-body}$_l([q_l, a_l])$
    \STATE $[z_{q_s}, z_{a_s}]\gets$\textit{VLM-body}$_s([q_s, a_s])$
    \STATE $[r_{q_s}, r_{a_l}]\gets\textit{\color{purple}{Recalibrator}}([z_{q_s}, z_{a_l}])$
    \STATE $\mathcal{L}_{ar}\gets \text{CE}(\textit{VLM-head}_l(r_{a_l}), gt_l)$
    \STATE $\mathcal{L}_{kl}\gets\mathcal{D}_{\text{KL}}(\textit{VLM-head}_l(z_{a_l}) \mid \textit{VLM-head}_l(r_{a_l}))$
    \STATE \textbf{Return:} $\mathcal{L}_{ar} + \mathcal{L}_{kl}$
\end{algorithmic}
\label{alg:1}
\end{algorithm}
}
{
\setlength{\textfloatsep}{0pt}
\begin{algorithm}[t!]
\caption{First stage Regularization (Alignment)}
\begin{algorithmic}[1]
    \STATE \textbf{Input:} ($q_l$, $a_l$), $gt_l$ (label index)
    \STATE $[z_{q_l}, z_{a_l}]\gets$\textit{VLM-body}$_l([q_l, a_l])$
    \STATE $[r_{q_l}, r_{a_l}]\gets\textit{\color{purple}{Recalibrator}}([z_{q_l}, z_{a_l}])$
    \STATE $\mathcal{L}_{ar}\gets \text{CE}(\textit{VLM-head}_l(r_{a_l}), gt_l)$
    \STATE $\mathcal{L}_{kl}\gets\mathcal{D}_{\text{KL}}(\textit{VLM-head}_l(z_{a_l}) \mid \textit{VLM-head}_l(r_{a_l}))$
    \STATE \textbf{Return:} $\mathcal{L}_{ar} + \mathcal{L}_{kl}$
\end{algorithmic}
\label{alg:2}
\end{algorithm}
}

\subsection{Design and Propagation of Recalibrator}
To construct a shared representation space, we first extract and align hidden features from both the large and small VLMs. Since the hidden space of a large VLM typically encompasses a richer vocabulary and more expressive representations, we regard it as the shared latent space which the small VLM is projected into. However, direct alignment of the two VLMs' tokens within this space is infeasible due to discrepancies in vocabulary size, token split, and token index ordering. To overcome this challenge, we draw inspiration from the principle of \textit{autoregressive modeling}, wherein a model predicts answer tokens for the given question tokens.

Specifically, we feed the same question-answer pair into both the large and small VLMs, obtaining their \textit{VLM-body} outputs as:
\begin{equation}
\begin{aligned}
[z_{q_l}, z_{a_l}] &= \textit{VLM-body}_l([q_l, a_l]), \quad&(l:\text{large})\\
[z_{q_s}, z_{a_s}] &= \textit{VLM-body}_s([q_s, a_s]), \quad&(l:\text{small})
\end{aligned}
\end{equation}
where $q$ and $a$ denote the question and answer tokens produced by each model’s vision encoder, vision projector, and word embedding~\cite{mikolov2013efficient}. We then extract the question features $z_{q_s}$ from the small VLM and the answer features $z_{a_l}$ from the large VLM, and concatenate them to form a joint sequence: $[z_{q_s}, z_{a_l}]$, which is then passed through Recalibrator. Because directly embedding both features into a shared space is infeasible due to different token types, we instead design an autoregressive loss that predicts the answer token index of the large VLM given the question token index of the small VLM. This loss encourages Recalibrator to effectively project the small VLM’s features into the large VLM’s latent space.

{
\setlength{\textfloatsep}{0pt}
\begin{algorithm}[t!]
\caption{Second stage Loss Functions (Distillation)}
\begin{algorithmic}[1]
    \STATE \textbf{Input:} ($q_l$, $a_l$), ($q_s$, $a_s$), ($gt_l$, $gt_s$) (label index)
    \STATE $[z_{q_l}, z_{a_l}]\gets$\textit{VLM-body}$_l([q_l, a_l])$
    \STATE ${[z_{q_s}, z_{a_s}]}\gets{\color{purple}{\textit{VLM-body}_s}}([q_s, a_s])$
    \STATE $[r_{q_s}, r_{a_l}]\gets\textit{Recalibrator}([z_{q_s}, z_{a_l}])$
    \STATE $\mathcal{L}_{ar}\gets\text{CE}(\textit{VLM-head}_l(r_{a_l}), gt_l)$
    \STATE $\mathcal{L}_{ar}\gets\mathcal{L}_{ar}+\text{CE}(\textit{VLM-head}_s(z_{a_s}), gt_s)$
    \STATE $\mathcal{L}_{kl}\gets\mathcal{D}_{\text{KL}}(\textit{VLM-head}_l(z_{a_l}) \mid \textit{VLM-head}_l(r_{a_l}))$
    \STATE \textbf{Return:} $\mathcal{L}_{ar} + \mathcal{L}_{kl}$
\end{algorithmic}
\label{alg:3}
\end{algorithm}
}

However, the hidden dimensions of large and small VLMs often differ. To address this, in Recalibrator, we first apply \textit{Proj-pre} to $z_{a_l}$, aligning its dimensionality to that of $z_{q_s}$. The concatenated sequence $[z_{q_s}, \textit{Proj-pre}(z_{a_l})]$ is then propagated through \textit{Rec-body} decoder blocks. The output of \textit{Rec-body} is subsequently passed through \textit{Proj-post} to restore the hidden dimensionality of the large VLM. This forward rule of Recalibrator is illustrated in \cref{fig:3}(c).

Finally, the answer part of the resulting features $[r_{q_s}, r_{a_l}]=\textit{Recalibrator}([z_{q_s}, z_{a_l}])$ is fed into the \textit{VLM-head} of the large VLM to compute the autoregressive loss:
\begin{equation}
\mathcal{L}_{ar} = \text{CE}(\textit{VLM-head}_l(r_{a_l}), gt_l),
\end{equation}
where $gt_l$ denotes the answer token index from the large VLM. Additionally, we employ a KL divergence loss to further enhance the Recalibrator’s projection capability in the first training stage and to distill knowledge from the teacher in the second training stage. This process enables the large VLM to interpret the hidden features of the small VLM, thus establishing a bridge for general-purpose distillation across heterogeneous token types.

\definecolor{colorful}{rgb}{0.9, 0.92, 0.850}
\newcommand{\cmark}{\ding{51}}%
\newcommand{\xmark}{\ding{55}}%
\begin{table}[t!]
\vspace{-3mm}
\centering
\caption{Evaluation of standard model size open-source VLMs and GenRecal on several challenging vision-language benchmarks: AI2D~\cite{kembhavi2016diagram}, ChartQA~\cite{masry2022chartqa}, MathVista~\cite{lu2023mathvista}, MMB~\cite{liu2023mmbench}, MMB$^{\text{CN}}$~\cite{liu2023mmbench}, MM-Vet~\cite{yu2023mm}, MMMU~\cite{yue2023mmmu}, MMMU-Pro~\cite{yue2024mmmu}, BLINK~\cite{fu2024blink}, SEED-2-Plus~\cite{li2024seed}, and RealWorldQA (RWQA). We use the notations of `\textit{XX}-GenRecal (\textit{YY})' where `\textit{XX}' and `\textit{YY}' denote the student (small) and teacher (large) VLM.}
\vspace{-3mm}
\label{tab:1}
\resizebox{\linewidth}{!}{
\renewcommand{\tabcolsep}{2mm}
\begin{tabular}{lccccccccccc}
\toprule
VLMs & AI2D & ChartQA & MathVista & MMB & MMB$^{\text{CN}}$ & MM-Vet & MMMU & MMMU-Pro & BLINK & SEED-2-Plus & RWQA\\
\midrule
Cambrian-1-8B~\cite{tong2024cambrian}
            & 73.0 
            & 73.3 
            & 49.0 
            & 75.9 
            & - 
            & 48.0 
            & 42.7 
            & -
            & 44.9
            & 59.7
            & 60.0\\ 
Cambrian-1-13B~\cite{tong2024cambrian}  
            & 73.6 
            & 73.8 
            & 48.0 
            & 75.7 
            & - 
            & 48.9 
            & 40.0 
            & -
            & 43.1
            & 60.0
            & 58.6\\ 
Eagle-8B~\cite{shi2024eagle}  
            & 76.1 
            & 80.1 
            & 52.7 
            & 75.9 
            & - 
            & 33.3 
            & 43.8 
            & -
            & 22.4
            & 57.5
            & 63.8\\ 
Eagle-13B~\cite{shi2024eagle}   
            & 74.0 
            & 77.6 
            & 54.4 
            & 75.7 
            & - 
            & 42.6 
            & 41.6 
            & -
            & 21.8
            & 60.2
            & 62.9\\ 
VILA1.5-8B~\cite{lin2023vila}  
            & 58.8    
            & -
            & 37.3
            & 75.3 
            & 69.9  
            & 43.2 
            & 38.6 
            & -
            & 39.5
            & 45.2
            & 43.4\\ 
VILA1.5-13B~\cite{lin2023vila}  
            & 69.9    
            & -
            & 42.5
            & 74.9 
            & 66.3 
            & 44.3 
            & 37.9 
            & -
            & 48.1
            & 50.2
            & 53.3\\ 
CogVLM2-19B~\cite{hong2024cogvlm2}
            & 73.4 
            & 81.0 
            & 38.6 
            & 80.5 
            & - 
            & 60.4 
            & 44.3 
            & -
            & -
            & 66.0
            & 62.9\\ 
LLaVA-OneVision-7B~\cite{li2024llava}
            & 81.4 
            & 80.0
            & 63.2  
            & 80.8 
            & - 
            & 57.5 
            & 48.8 
            & 24.1
            & 53.0
            & 65.4
            & 69.9\\ 
InternVL2-8B~\cite{chen2023internvl}
            & 83.8 
            & 83.3 
            & 58.3  
            & 81.7 
            & 81.2 
            & 54.2 
            & 49.3 
            & 29.0
            & 50.9
            & 67.3
            & 64.2\\ 
MiniCPM-V2.5-8B~\cite{yao2024minicpm}
            & 78.4 
            & - 
            & 54.3  
            & 77.2 
            & 74.2 
            & 52.8 
            & 45.8 
            & -
            & -
            & 61.4
            & 63.5\\ 
MiniCPM-V2.6-8B~\cite{yao2024minicpm}
            & 82.1 
            & -
            & 60.6  
            & - 
            & - 
            & 60.0 
            & 49.8 
            & 27.2
            & 55.2
            & 65.7
            & 65.0\\ 
Qwen2-VL-7B~\cite{wang2024qwen2vl}
            & 77.5 
            & 83.0
            & 58.2 
            & 83.0 
            & 80.5 
            & 62.0 
            & 54.1 
            & 30.5
            & 53.8
            & 68.6
            & 68.5\\ 
InternVL2.5-8B~\cite{chen2024expanding}
            & 84.8 
            & 84.8
            & 64.4 
            & 84.6  
            & 82.6 
            & 62.8 
            & 56.0 
            & 34.3 
            & 54.8
            & 69.7
            & 70.1\\ 
\midrule
\rowcolor{colorful}
Qwen2-VL-7B-GenRecal (InternVL2.5-78B)
            & \textbf{93.9} 
            & \textbf{95.3} 
            & 68.8 
            & 88.4 
            & 87.4 
            & 70.4 
            & 65.6 
            & \textbf{49.6}
            & 64.3
            & 70.7
            & 79.2\\ 
\rowcolor{colorful}
InternVL2.5-8B-GenRecal (InternVL2.5-78B)
            & 93.0 
            & 93.6 
            & \textbf{74.9} 
            & \textbf{89.5} 
            & \textbf{88.2} 
            & \textbf{73.2} 
            & \textbf{68.1} 
            & 48.8
            & \textbf{65.3}
            & \textbf{72.3} 
            & \textbf{81.4}\\ 
\midrule
VILA1.5-3B~\cite{lin2023vila}
            & 57.9 
            & - 
            & 31.6 
            & - 
            & - 
            & 38.8 
            & 34.2 
            & - 
            & 39.7
            & 41.4 
            & 53.2 \\ 
Phi-3.5-Vision-4B~\cite{abdin2024phi}
            & 77.8
            & 81.8 
            & 43.9 
            & 76.0 
            & 66.1 
            & 43.2 
            & 43.0 
            & 19.7 
            & 58.3
            & 62.2
            & 53.6\\ 
InternVL2-4B~\cite{chen2023internvl}
            & 78.9 
            & 81.5 
            & 58.6 
            & 78.6 
            & 73.9 
            & 51.0 
            & 34.3 
            & 32.7 
            & 46.1
            & 63.9
            & 60.7\\ 
InternVL2.5-4B~\cite{chen2023internvl}
            & 81.4 
            & 84.0  
            & 60.5 
            & 81.1 
            & 79.3 
            & 60.6 
            & 52.3  
            & 32.7
            & 50.8 
            & 66.9
            & 64.3\\ 
\midrule
\rowcolor{colorful}
InternVL2.5-4B-GenRecal (InternVL2.5-78B)  
            & \textbf{90.0}
            & \textbf{91.1} 
            & \textbf{70.4}
            & \textbf{88.4}
            & \textbf{86.2}
            & \textbf{66.1}
            & \textbf{58.3}
            & \textbf{45.9}
            & \textbf{63.5}
            & \textbf{69.3}
            & \textbf{75.2}\\ 
\midrule
InternVL2-2B~\cite{chen2023internvl}
            & 74.1 
            & 76.2 
            & 46.3  
            & 73.2 
            & 70.9 
            & 39.5 
            & 34.3 
            & 18.2
            & 43.8
            & 60.0
            & 57.3\\ 
Qwen2-VL-2B~\cite{wang2024qwen2vl}
            & 60.2 
            & 73.5
            & 43.0 
            & 74.9 
            & 73.5 
            & 49.5 
            & 41.1 
            & 21.2
            & 45.2
            & 61.2
            & 62.6\\ 
Aquila-VL-2B~\cite{chen2024expanding}
            & 75.0
            & 76.5
            & 59.0
            & -
            & -
            & 43.8
            & 47.4
            & -
            & 34.1
            & 63.0
            & 65.0\\ 
InternVL2.5-2B~\cite{chen2024expanding}
            & 74.9 
            & 79.2 
            & 51.3 
            & 74.7 
            & 71.9 
            & 60.8 
            & 43.6  
            & 23.7 
            & 44.0 
            & 60.0
            & 60.1\\ 
\midrule
\rowcolor{colorful}
Qwen2-VL-2B-GenRecal (InternVL2.5-78B) 
            &  89.1
            &  \textbf{90.9}
            &  60.5
            &  82.9
            &  81.0
            &  57.3
            &  52.9
            & \textbf{37.5}
            & 53.4
            & 65.5
            & 72.8\\ 
\rowcolor{colorful}
InternVL2.5-2B-GenRecal (InternVL2.5-78B) 
            & \textbf{91.8} 
            &  90.7
            & \textbf{62.5} 
            & \textbf{84.1} 
            & \textbf{80.4} 
            & \textbf{61.2} 
            & \textbf{59.2} 
            & 36.6
            & \textbf{55.5}
            & \textbf{67.3}
            & \textbf{73.0}\\ 
\midrule
LLaVA-OneVision-0.5B~\cite{li2024llava}
            & 57.1
            & 61.4
            & 34.8
            & 61.6
            & 55.5
            & 32.2
            & 31.4
            & -
            & 52.1
            & 45.7
            & 55.6\\ 
InternVL2-1B~\cite{chen2023internvl} 
            & 64.1
            & 72.9
            & 37.7
            & 65.4
            & 60.7
            & 32.7
            & 36.7
            & 14.8
            & 38.6
            & 54.3
            & 50.3\\ 
InternVL2.5-1B~\cite{chen2024expanding} 
            & 69.3
            & 75.9
            & 43.2
            & 70.7
            & 66.3
            & 48.8
            & 40.9
            & 19.4 
            & 42.0 
            & 59.0 
            & 57.5\\ 
\midrule
\rowcolor{colorful}
InternVL2.5-1B-GenRecal (InternVL2.5-78B)  
            & \textbf{82.9}
            & \textbf{89.4}
            & \textbf{56.5}
            & \textbf{80.1}
            & \textbf{74.4}
            & \textbf{54.4}
            & \textbf{45.6}
            & \textbf{28.7}
            & \textbf{48.5}
            & \textbf{65.1}
            & \textbf{70.8}\\ 
\bottomrule
\end{tabular}
}
\vspace{-5mm}
\end{table}

\subsection{Training Process}
To achieve general-purpose distillation, we adopt a three-stage training process. In the first stage, only Recalibrator is trained while keeping all parameters of both the large and small VLMs frozen, as illustrated in \cref{fig:3}(a). During this stage, we compute both the autoregressive loss and the KL divergence, described in \cref{alg:1} where the {\color{purple}{purple}} components indicate the trainable parameters. This step primarily serves to align the feature representations between the large and small VLMs. For the first stage, we introduce regularization loss, as illustrated in \cref{fig:4}, which is designed to prevent the feature representations of Recalibrator from drifting too far from those of the large VLM. We use both \cref{alg:1} and \cref{alg:2} together in the first stage.

In the second stage, we keep using the same loss functions of the first stage and additionally incorporate the small VLM’s own autoregressive loss in \cref{alg:3}. As depicted in \cref{fig:3}(b), this stage enables knowledge distillation from the large VLM to the small one by training the small VLM’s \textit{VLM-body}. In the final stage, we remove both Recalibrator and the large VLM, and fine-tune the student VLM. Specifically, we train all parameters of the student VLM except for the vision encoder through supervised fine-tuning (SFT) to further enhance its instruction-following capability.
\section{Experiments}
\label{sec:experiment}

\begin{table}[t!]
\vspace{-3mm}
\centering
\caption{(b) Comparison of GenRecal with large-scale open-source and closed-source VLMs on challenging benchmarks: MMB~\cite{liu2023mmbench}, MM-Vet~\cite{yu2023mm}, MM-Vet-v2~\cite{yu2024mm}, MMMU~\cite{yue2023mmmu}, MMMU-Pro~\cite{yue2024mmmu}, MMStar~\cite{chen2024we}, AI2D~\cite{kembhavi2016diagram}, ChartQA~\cite{masry2022chartqa}, SEED-2-Plus~\cite{li2024seed}, MathVista~\cite{lu2023mathvista}, BLINK~\cite{fu2024blink}, RealWorldQA (RWQA).}
\vspace{-3mm}
\label{tab:2}
\resizebox{\linewidth}{!}{
\renewcommand{\tabcolsep}{1mm}
\begin{tabular}{lcccccccccccc}
\toprule
VLMs & MMB & MM-Vet & MM-Vet-v2 & MMMU & MMMU-Pro & MMStar & AI2D & ChartQA & SEED-2-Plus & MathVista & BLINK & RWQA \\
\midrule
NVLM-72B~\cite{nvlm2024}
& -
& 58.9
& -
& 59.7
& -
& 63.7 
& 85.2
& 86.0
& 68.4
& 66.6
& 48.0
& 69.9\\ 
LLaVA-OneVision-72B~\cite{li2024llava}
& 85.8
& 60.6
& -
& 56.8
& 31.0
& 65.8
& 85.6
& 83.7
& -
& 67.5
& 55.4
& 71.9\\ 
Molmo-72B~\cite{deitke2024molmo}
& -
& 61.1
& -
& 54.1
& -
& 63.3
& 83.4
& 87.3
& -
& 58.6
& -
& 73.7\\ 
Qwen2-VL-72B~\cite{wang2024qwen2vl}
& 86.5
& 74.0
& 68.7
& 64.5
& 46.2
& 68.3
& 88.1
& 88.3
& \textbf{72.3}
& 70.5
& 60.5
& 77.8\\ 
InternVL2-76B~\cite{chen2023internvl}
& 86.5
& 65.7
& 68.4
& 62.7
& 40.0
& 67.4 
& 87.6
& 88.4
& 69.7
& 65.5
& 56.8
& 72.2\\ 
InternVL2.5-78B~\cite{li2024llava}
& 88.3
& 72.3 
& 65.5
& \textbf{70.1}
& 48.6 
& \textbf{69.5} 
& 89.1
& 88.3 
& 71.3
& 72.3 
& 63.8
& 78.7\\ 
Claude-3.5-Sonnet~\cite{claude3series2024}
&  82.6
&  70.1
& \textbf{71.8}
& 68.3
& 51.5 
& 65.1
& 81.2
& 90.8
& 71.7
& 67.7
& 60.1
& 60.1\\ 
Gemini-1.5-Pro~\cite{team2023gemini}
& 73.9
&  64.0
& 66.9
& 62.2
& 46.9
& 59.1 
& 79.1
&  87.2
& 70.8
& 63.9
& 59.1
& 67.5\\ 
GPT-4o (0513)~\cite{gptsyscard}
&  83.4
&  69.1
&  71.0
& 69.1
& \textbf{51.9} 
&  64.7
& 84.6
& 85.7
& 72.0
& 63.8
& \textbf{68.0}
& 75.4\\ 
\midrule
\rowcolor{colorful}
InternVL2.5-8B-GenRecal (NVLM-72B) 
& 88.8
& 63.9
& 59.3
& 60.3
& 34.7
& 65.1
& 88.5
& 84.3
& 69.3
& 67.5
& 55.6
& 72.5\\ 
\rowcolor{colorful}
InternVL2.5-8B-GenRecal (InternVL2-76B) 
& 89.0
& 72.4
& 65.0
& 64.6
& 42.8
& 65.8
& 92.1
& 90.4
& 70.9
& 71.9
& 57.7
& 75.5\\ 
\rowcolor{colorful}
InternVL2.5-8B-GenRecal (Qwen2-VL-72B) 
& \textbf{89.5}
& 71.4
& 62.6
& 65.6
& 43.8
& 66.1
& 92.2
& 87.5
& 71.3
& 71.4
& 61.4
& 78.8\\ 
\rowcolor{colorful}
InternVL2.5-8B-GenRecal (InternVL2.5-78B)
& \textbf{89.5}
& \textbf{73.2}
& 67.2
& 68.1
& 48.8
& 67.8
& \textbf{93.0}
& \textbf{93.6}
& \textbf{72.3}
& \textbf{74.9}
& 65.3
& \textbf{81.4}\\ 
\bottomrule
\end{tabular}
}
\vspace{-5mm}
\end{table}


\subsection{Implementation Detail}
To ensure reproducibility, we outline three key technical aspects of GenRecal: (a) the structure of the Recalibrator, (b) the projection layers, and (c) training and evaluation details.

\vspace{1.5mm}
\noindent\textbf{(a) Structure Detail of Recalibrator.}
This is composed of two transformer decoder blocks (\textit{Rec-body}) and two projectors (\textit{Proj-pre} and \textit{Proj-post}). The decoder blocks follow all configurations of the small VLM's decoder block, such as the causal mask, hidden dimensions, number of heads, and FFN structure. We use transformer decoder blocks for \textit{Rec-body} because they naturally process the concatenated student and teacher features in \cref{fig:3} as a sequence. Moreover, these features must preserve token ordering, so we introduce new positional embeddings (NPE). This design enables effective sequential modeling and representation alignment between heterogeneous VLMs. To this end, we employ an additional RoPE~\cite{su2024roformer} to realign their positional embeddings, and their position IDs are reassigned accordingly. Lastly, we apply an additional layer norm~\cite{ba2016layernormalization} to the output features of the Recalibrator for stable adaptation.

\vspace{1.5mm}
\noindent\textbf{(b) Projection Layer Design.}
Two projection layers (\textit{Proj-pre} and \textit{Proj-post}) are linear layers introduced for hidden dimension matching. Similar to the role of vision projector that matches the hidden dimensions between a CLIP-like vision encoder and an LLM, \textit{Proj-pre} matches the teacher features to the student hidden dimension, while \textit{Proj-post} restores the dimension back to the teacher hidden dimension so that the large \textit{VLM-head} can be utilized for distillation. This explanation is technically depicted in \cref{fig:3}(c) and \cref{fig:4}.

\begin{figure}[t!]
    \vspace{-3mm}
    \centering
    \includegraphics[width=0.4\textwidth]{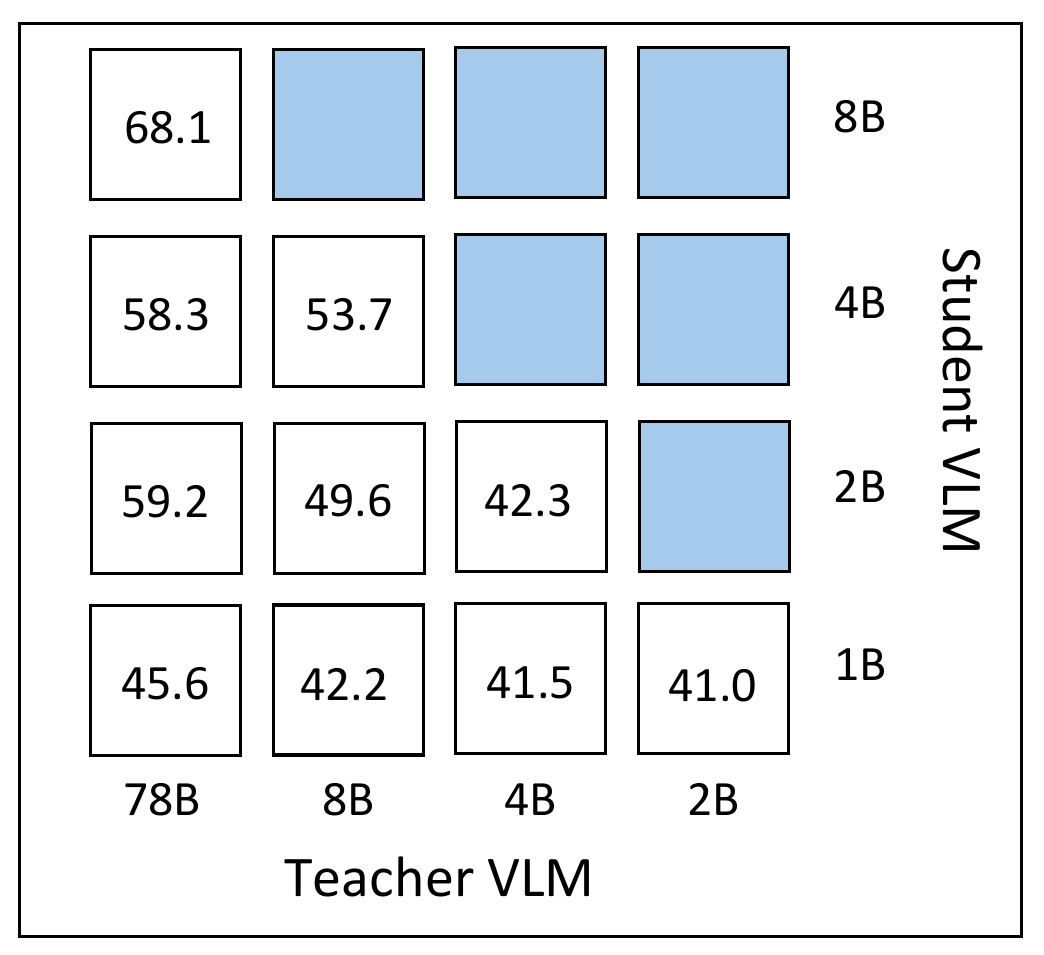}
    \vspace{-3mm}
    \caption{Distillation performance on MMMU~\cite{yue2023mmmu} for various pairings os teacher and student VLM. Each cell indicates the resulting score when using the corresponding teacher (rows) and student (columns) model sizes.}
    \label{fig:5}
    \vspace{-5mm}
\end{figure}

\vspace{1.5mm}
\noindent\textbf{(c) Details of Training and Evaluation.}
By using DeepSpeed engine with ZeRO-3~\cite{rajbhandari2020zero}, we train and evaluate GenRecal on 128 NVIDIA A100 80GB GPUs. We use AdamW optimizer~\cite{loshchilov2018decoupled} and apply a linearly decayed learning rate from 1e-4 to 1e-5 at each training stage. The first and second stages use the entire 9M dataset (see \appref{app:E} for dataset composition and analysis), taking approximately 2--4 hours and 8--11 hours, respectively, depending on model size. The last stage takes 4--6 hours on a 6M dataset obtained by removing general visual question answering samples~\cite{li2024llava}. For stable training, we handle large batch sizes via gradient accumulation with 16 steps. We use 4 (first and third stages) or 2 (second stage) samples per GPU, leading to a total batch size of 8192 (128$\times$16$\times$4) or 4096 (128$\times$16$\times$2). For evaluation, we remove the Recalibrator and the large VLM, and let the small VLM generate answers using the default generation hyperparameters.

\vspace{1.5mm}
\noindent\textbf{(d) Dataset Composition.}
Since the 9M corpus has a decisive impact on the results, we summarize it here. It covers general visual question answering, dense captioning, chart/diagram/document understanding, commonsense knowledge, science \& math, and multi-dimensional reasoning, curated from public sources such as LLaVA-OneVision~\cite{li2024llava}, Cambrian~\cite{tong2024cambrian}, Infinity-MM~\cite{gu2024infinity}, and DenseFusion~\cite{li2024densefusion}, among others. For analysis we group these into three domains---\textit{Knowledge}, \textit{Science \& Math}, and \textit{Chart \& Document}---which align with benchmarks such as MMMU~\cite{yue2023mmmu}, MathVista~\cite{lu2023mathvista}, and ChartQA~\cite{masry2022chartqa}, respectively. A full source list and per-domain breakdown are provided in \appref{app:E}.

\subsection{Validation and Analysis for GenRecal}

\begin{figure}[t!]
    \vspace{-3mm}
    \centering
    \includegraphics[width=\textwidth]{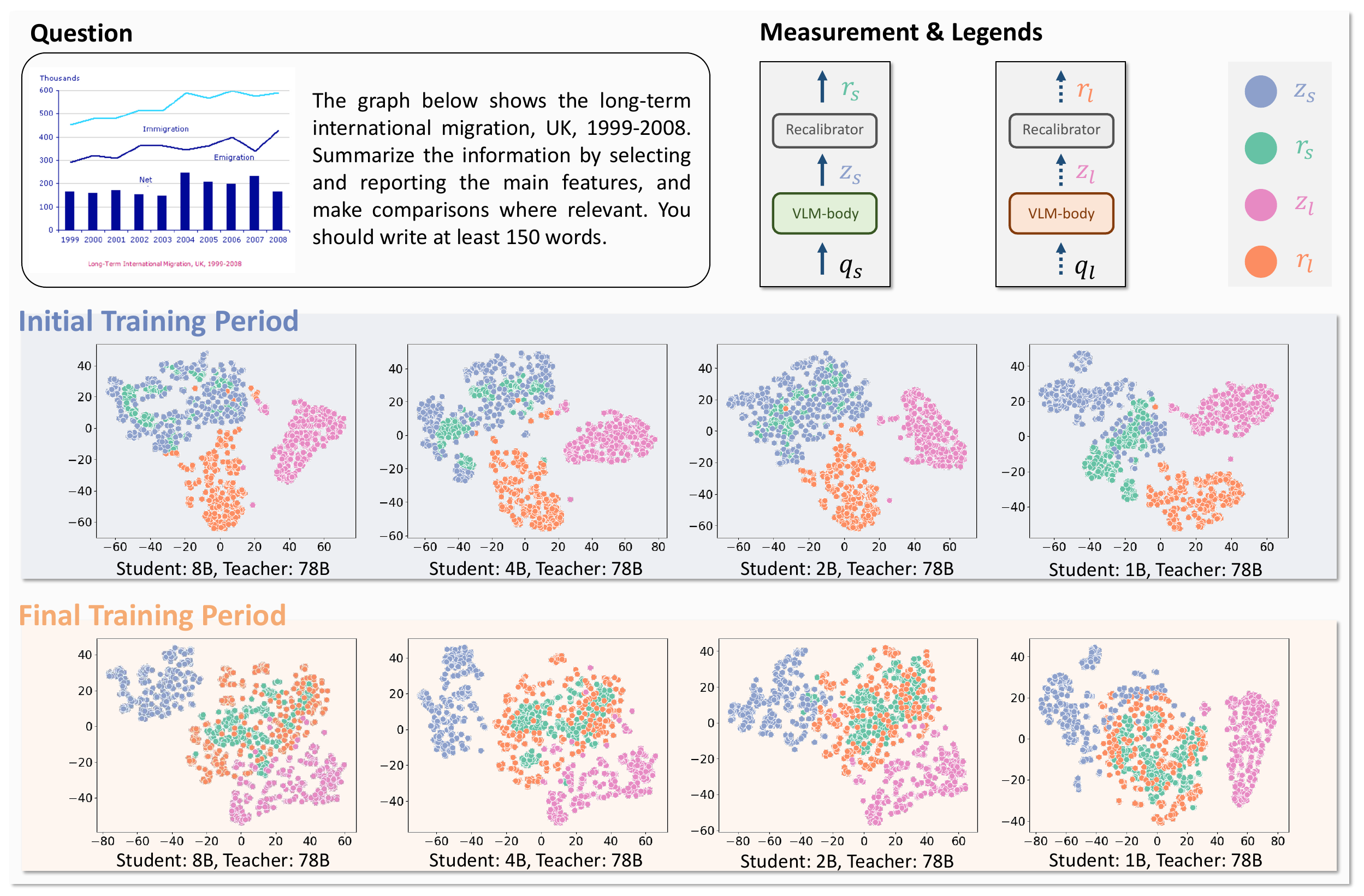}
    \vspace{-7mm}
    \caption{An overview of our training pipeline, illustrating both the question prompt and the measurement/legend annotations (top), followed by t-SNE visualizations (bottom) of teacher and student VLM pairings at the initial and final training stages. The question prompt (upper-left) shows the format of the question, while the measurement and legend box (upper-right) shows key model components to measure. Each scatter plot in the lower panels corresponds to a different combination of teacher and student VLM sizes, capturing how the learned representations evolve from early to later training.}
    \label{fig:6}
    \vspace{-5mm}
\end{figure}

\noindent\textbf{Higher Size, Higher Performance.}
\cref{tab:1} demonstrates that GenRecal outperforms not only standard-size VLMs but also smaller ones, highlighting that GenRecal generalizes well across model sizes. This table also shows that choosing a higher-performing small VLM such as InternVL2.5-8B~\cite{chen2024expanding} substantially improves the overall distillation results. Furthermore, for a fixed large VLM, selecting a smaller student VLM leads to lower distillation performance, implying that employing more capable small VLMs is crucial for achieving higher distillation performance.

Next, we examine the effect of changing large VLMs to determine whether the aforementioned property holds consistently across different large VLMs. As shown in \cref{fig:2} and \cref{tab:2}, selecting more powerful large VLMs leads to greater performance improvements. To investigate this further, we use the InternVL2.5 series~\cite{chen2024expanding} and experiment with various combinations of large and small VLM sizes, as illustrated in \cref{fig:5}. The results consistently indicate that choosing larger teacher and student VLMs leads to higher distillation performance.

\vspace{1.5mm}
\noindent\textbf{Analysis of Recalibrator.}
We now focus on the Recalibrator to explore its role in general-purpose distillation. First, we show that the loss curves for training the Recalibrator converge stably across various combinations of large and small VLMs (see \appref{app:D}). Since the Recalibrator losses (`Recalib$(q_s, a_l)$' and `Recalib$(q_l, a_l)$') are minimized well relative to the perplexity of `SmallVLM$(q_s, a_s)$' and `LargeVLM$(q_l, a_l)$', we infer that the Recalibrator successfully aligns the feature representations of large and small VLMs, and that it generalizes well across teacher--student combinations. Furthermore, \cref{fig:6} visualizes multiple feature spaces to assess whether the large- and small-VLM features after the Recalibrator are truly matched and can be regarded as a shared representation. The small-VLM features before the Recalibrator are initially close to those after it, but by the end of training they become matched to the large-VLM features after the Recalibrator. This indicates that the Recalibrator is effective at forming a shared feature representation for large and small VLMs.

\begin{table}[t!]
\vspace{-3mm}
\centering
\begin{minipage}[t]{0.48\linewidth}
\centering
\caption{Comparing the final distillation performances from the teacher model: InternVL2.5-78B~\cite{chen2024expanding} with and without the regularization term (denoted as ``Reg''). Note that, the student models are each InternVL2.5-8B, 4B, 2B, and 1B.}
\label{tab:3}
\vspace{-3mm}
\resizebox{\linewidth}{!}{
\renewcommand{\tabcolsep}{5.3mm}
\begin{tabular}{lccccc}
\toprule
VLMs         &Reg         & MMB  & MathVista & MM-Vet & MMMU\\
\midrule
InternVL2.5-8B&-          & 84.6 & 64.4  & 62.8  & 56.0\\
\cdashline{1-6}\noalign{\vskip 0.5ex}
w. GenRecal      &\xmark      & 88.2 & 69.8  & 63.5  & 58.9\\
\rowcolor{colorful}
w. GenRecal      &\cmark      & \textbf{89.5} & \textbf{74.9} & \textbf{73.2} &\textbf{68.1}\\
\midrule
InternVL2.5-4B&-          & 81.1  & 60.5 & 60.6 & 52.3\\
\cdashline{1-6}\noalign{\vskip 0.5ex}
w. GenRecal      &\xmark      & 82.9  & 61.7 & 61.1 & 53.6\\
\rowcolor{colorful}
w. GenRecal      &\cmark      & \textbf{88.4} & \textbf{70.4} & \textbf{66.1} &\textbf{58.3}\\
\midrule
InternVL2.5-2B&-          & 74.7  & 51.3 & 60.8 & 43.6\\
\cdashline{1-6}\noalign{\vskip 0.5ex}
w. GenRecal      &\xmark      & 75.8 & 	53.2 & 60.9 & 45.4\\
\rowcolor{colorful}
w. GenRecal      &\cmark      & \textbf{84.1} & \textbf{62.5} & \textbf{61.2} &\textbf{59.2}\\
\midrule
InternVL2.5-1B&-          & 70.7  & 43.2 & 48.8 & 40.9\\
\cdashline{1-6}\noalign{\vskip 0.5ex}
w. GenRecal      &\xmark      & 72.6  & 45.9 & 49.8 & 41.1\\
\rowcolor{colorful}
w. GenRecal      &\cmark      & \textbf{80.1} & \textbf{56.5} & \textbf{54.4} &\textbf{45.6}\\
\bottomrule
\end{tabular}
}

\end{minipage}
\hspace{0.01\linewidth}
\begin{minipage}[t]{0.48\linewidth}
\centering
\caption{Comparing SFT, traditional distillations, and GenRecal where we use equal training dataset (Section~\ref{sec:experiment}.1(b)) and choose the teacher (Qwen2-VL-72B~\cite{wang2024qwen2vl}) that has the same token type of student.}
\label{tab:4}
\vspace{-3mm}
\resizebox{\linewidth}{!}{
\renewcommand{\tabcolsep}{6mm}
\begin{tabular}{lcccc}
\toprule
VLMs               & MMB   & MathVista & MM-Vet & MMMU\\
\midrule
Qwen2-VL-7B        & 83.0  & 58.2      & 62.0   & 54.1 \\
\cdashline{1-5}\noalign{\vskip 0.5ex}
w. SFT             & 84.3  & 60.5      & 64.2   & 56.3 \\
w. MiniLLM         & 84.4  & 61.3      & 65.1   & 57.4 \\
w. DistiLLM        & 84.8  & 61.5      & 65.3   & 57.9 \\
w. LLaVA-KD        & 85.0  & 61.8      & 65.9   & 58.2 \\
\rowcolor{colorful}
w. GenRecal            & \textbf{87.8}  & \textbf{69.5}      & \textbf{67.8}   & \textbf{64.2}\\
\midrule
Qwen2-VL-2B        & 74.9  & 43.0      & 49.5   & 41.1 \\
\cdashline{1-5}\noalign{\vskip 0.5ex}
w. SFT             & 76.3  & 45.8      & 51.8   & 44.3 \\
w. MiniLLM         & 77.5  & 46.2      & 52.5   & 45.3 \\
w. DistiLLM        & 77.9  & 46.6      & 53.0   & 46.1 \\
w. LLaVA-KD        & 78.3  & 46.9      & 53.7   & 46.9 \\
\rowcolor{colorful}
w. GenRecal            & \textbf{82.9}  & \textbf{58.3}      & \textbf{57.1}   & \textbf{51.4}\\
\bottomrule
\end{tabular}
}
\end{minipage}
\end{table}
\vspace{-5mm}


\begin{figure}[t!]
    \vspace{-3mm}
    \centering
    \includegraphics[width=\textwidth]{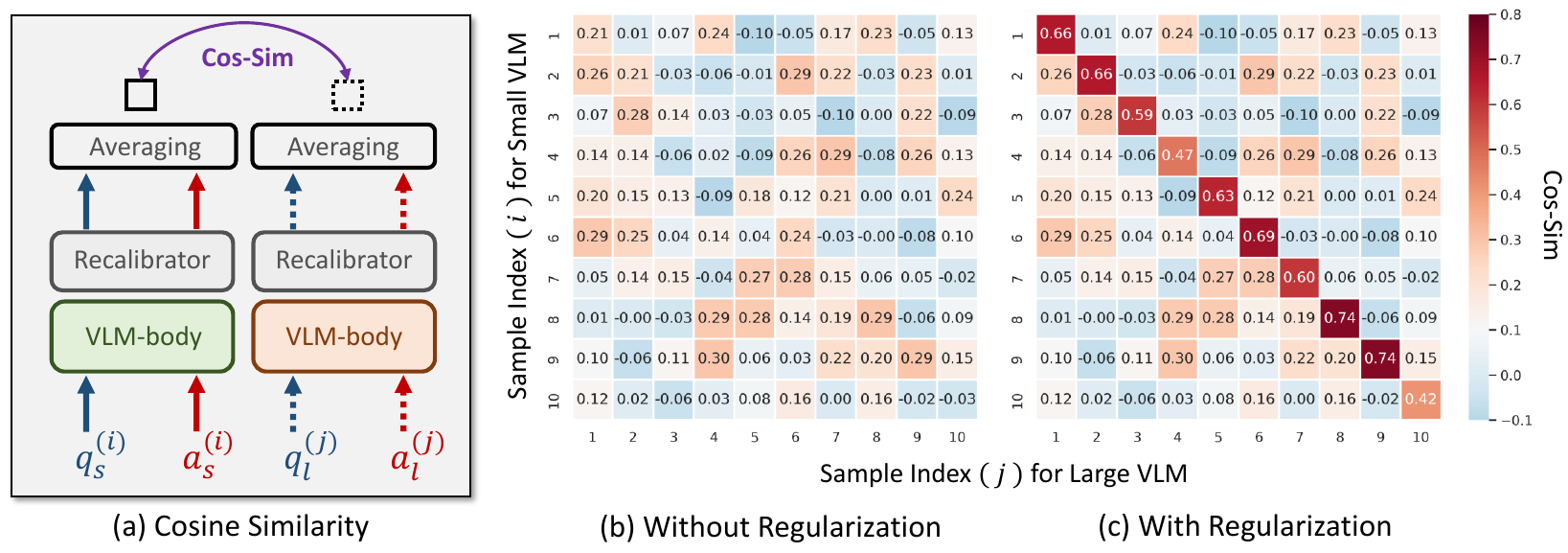}
    \vspace{-7mm}
    \caption{The process of (a) computing cosine similarity, (b) the result without regularization or (c) with regularization. Because the number of the embedded tokens for small and large VLMs are naturally different due to vocabulary size, token split, and token ordering, therefore we do average the output tokens and compute cosine similarity.}
    \label{fig:7}
    \vspace{-5mm}
\end{figure}

\vspace{6mm}
\noindent\textbf{Importance of Regularization.}
We conduct an ablation study that investigates how important the regularization term is for the final distillation performances. We first remove the regularization term and analyze the differences before and after its removal. To illustrate this in detail, we simply prepare 10 different question-answer pairs and input them through \textit{VLM-body} of both small and large VLMs. Afterwards, we propagate their output features into Recalibrator, as described in \cref{fig:7}(a). We then compute a $10 \times 10$ cosine similarity matrix between Recalibrator’s output features from the small and large VLMs. To ensure reliable cosine similarity values, we use all training dataset samples and compute the average of the cosine similarity values for all the samples.

With the regularization term included, as shown in \cref{fig:7}(c), we observe that the diagonal values in the cosine similarity matrix are significantly higher than the off-diagonal values. This implies that the features of small and large VLMs are explicitly shared through Recalibrator, aligned with the results in \cref{fig:6} as well. In contrast, when the regularization term is removed, as depicted in \cref{fig:7}(b), the diagonal values are comparable with the off-diagonal values. This indicates that Recalibrator fails to explicitly align large and small VLMs' features, preventing feature matching and sharing. As a result, this leads to a significant drop in final distillation performances, as reported in \cref{tab:3}. Insufficient alignment of large and student VLMs reduces efficient knowledge transfer during distillation. These findings highlight the crucial role of the regularization term in feature alignment and knowledge transfer for general-purpose distillation.

\vspace{1.5mm}
\noindent\textbf{Autoregressive Loss for Representation Alignment.}
To align the student representations with the teacher’s latent space, GenRecal uses two objectives: KL divergence loss ($\mathcal{L}_{kl}$) and autoregressive loss ($\mathcal{L}_{ar}$). The KL divergence loss performs standard distillation by matching the logit-level distribution of teacher and student. However, matching distributions alone does not guarantee precise alignment at the token level. The autoregressive loss provides explicit hard-target supervision, allowing it to predict the exact ground-truth token sequence. In short, $\mathcal{L}_{kl}$ transfers knowledge at the distribution level, whereas $\mathcal{L}_{ar}$ ensures exact alignment at the token level. As shown in \cref{tab:arloss}, removing $\mathcal{L}_{ar}$ consistently degrades performance across benchmarks, confirming that autoregressive supervision is essential for effective representation alignment.

\begin{table}[t!]
\vspace{-3mm}
\caption{Importance of the autoregressive loss ($\mathcal{L}_{ar}$) for representation alignment. We compare the baseline VLMs, GenRecal trained without $\mathcal{L}_{ar}$ (\xmark, using only $\mathcal{L}_{kl}$), and GenRecal trained with $\mathcal{L}_{ar}$ (\cmark), using InternVL2.5-78B~\cite{chen2024expanding} as the teacher VLM.}
\label{tab:arloss}
\vspace{-3mm}
\centering
\small
\resizebox{\linewidth}{!}{
\renewcommand{\tabcolsep}{1mm}
\begin{tabular}{lccccccccccccc}
\toprule
VLMs &
$\mathcal{L}_{ar}$&
AI2D &
ChartQA &
MathVista &
MMB &
MMB$^{\text{CN}}$ &
MM-Vet &
MMMU &
MMMU-Pro &
BLINK &
SEED-2-Plus &
RWQA &
Avg \\
\midrule
Qwen2-VL-7B~\cite{wang2024qwen2vl}
& -
& 77.5 & 83.0 & 58.2 & 83.0 & 80.5 & 62.0 & 54.1 & 30.5 & 53.8 & 68.6 & 68.5 & 65.4 \\
InternVL2.5-8B~\cite{chen2024expanding}
& -
& 84.8 & 84.8 & 64.4 & 84.6 & 82.6 & 62.8 & 56.0 & 34.3 & 54.8 & 69.7 & 70.1 & 68.1 \\
\midrule
Qwen2-VL-7B-GenRecal (InternVL2.5-78B)
& \xmark
& 66.2 & 67.5 & 47.8  & 63.4 & 55.9 & 49.6 & 45.8 & 30.5 & 52.1 & 65.3 & 60.9 & 55.0 \\
InternVL2.5-8B-GenRecal (InternVL2.5-78B)
& \xmark
& 68.4 & 70.1 & 55.2 & 68.9 & 62.5 & 55.4 & 50.6 & 35.1 & 58.7 & 68.2 & 64.4 & 59.8 \\
\midrule
Qwen2-VL-7B-GenRecal (InternVL2.5-78B)
& \cmark
& 93.9 & 95.3 & 68.8 & 88.4 & 87.4 & 70.4 & 65.6 & 49.6 & 64.3 & 70.7 & 79.2 & 75.8 \\
InternVL2.5-8B-GenRecal (InternVL2.5-78B)
& \cmark
& 93.0 & 93.6 & 74.9 & 89.5 & 88.2 & 73.2 & 68.1 & 48.8 & 65.3 & 72.3 & 81.4 & 77.1 \\
\bottomrule
\end{tabular}
}
\vspace{-4mm}
\end{table}


\subsection{Ablation Studies: Training Computation and Configurations}
In \cref{tab:5}(a), we compare the model sizes and training FLOPs of the small VLM, large VLM, and Recalibrator to assess whether Recalibrator introduces additional computational overhead during training. The results show that Recalibrator’s FLOPs are substantially lower than those of both the large and small VLMs. Note that Recalibrator is removed at inference time, thus incurring no additional computational cost during inference. We further conduct a series of ablation studies to examine various configurations affecting the distillation performance of GenRecal.
As shown in \cref{tab:5}(b), we vary the decoder depth of Recalibrator from 1 to 20 and select two depths for layer number of \textit{Rec-body} based on a trade-off between computational efficiency and performance. Additionally, \cref{tab:5}(c) demonstrates the importance of employing a new positional embedding (NPE) to handle newly introduced position IDs and to integrate RoPE~\cite{su2024roformer} from both large and small VLMs, which have different token types.

To analyze the effect of dataset scale, we control the dataset size from 1M to 9M samples, as shown in \cref{tab:5}(d). The performance improvements between 5M and 9M samples are marginal, suggesting that a 5M dataset is a practical choice for users with limited training resources. Furthermore, \cref{tab:5}(e) shows that fine-tuning the large VLM on our dataset yields further gains, indicating that the small VLM trained under the same setting has the potential to match or even surpass the large VLM’s performance.

We also compare GenRecal with recent cross-tokenizer distillation methods in \cref{tab:5}(f). These baselines perform distillation via cross-token matching in the logit space using the classical Wasserstein distance~\cite{boizard2025towards} or optimal transport~\cite{cui2025multi}, whereas GenRecal introduces the Recalibrator to align hidden features for semantic correspondence. GenRecal achieves a substantial improvement over these approaches, as discussed in the next section. Lastly, \cref{tab:5}(g)–(i) demonstrate GenRecal’s strong compatibility across different LLMs, VLM families, and recently released VLMs.
In all cases, GenRecal consistently delivers superior performance, validating its general applicability to diverse model architectures.

\begin{table}[t!]
\centering
\vspace{-3mm}
\caption{FLOPs comparison and several ablation studies on various configurations influencing the performance of GenRecal. Note that the teacher VLM in (b), (c), (d), (e), and (f) is InternVL2.5-78B~\cite{chen2024expanding}, and the student VLM in (b) is InternVL2.5-8B~\cite{chen2024expanding}.}
\label{tab:5}
\centering
\vspace{-5mm}
\begin{minipage}[t]{0.32\linewidth}
\caption*{\scriptsize (a) Model Size and FLOPs}
\vspace{-3mm}
\centering
\resizebox{\linewidth}{!}{
\renewcommand{\tabcolsep}{1mm}
\begin{tabular}{lll}
\toprule
Small VLM & Large VLM & Recalibrator Size \\
\midrule
Qwen2-VL-7B ($25.7 \times 10^{12}$) & Qwen2-VL-72B ($260 \times 10^{12}$) & 524.8M ($2.27 \times 10^{12}$) \\
\cdashline{1-3}\noalign{\vskip 0.5ex}
InternVL2.5-8B ($27.1 \times 10^{12}$) & InternVL2.5-78B ($269 \times 10^{12}$) & 503.3M ($2.2 \times 10^{12}$) \\
\cdashline{1-3}\noalign{\vskip 0.5ex}
InternVL2.5-8B ($27.1 \times 10^{12}$) & Qwen2-VL-72B ($260 \times 10^{12}$) & 503.3M ($2.2 \times 10^{12}$) \\
\cdashline{1-3}\noalign{\vskip 0.5ex}
Qwen2-VL-7B ($25.7 \times 10^{12}$) & InternVL2.5-78B ($269 \times 10^{12}$) & 524.8M ($2.27 \times 10^{12}$) \\
\bottomrule
\end{tabular}
}

\caption*{\scriptsize (d) Training Dataset Size}
\vspace{-3mm}
\centering
\resizebox{\linewidth}{!}{
\renewcommand{\tabcolsep}{5mm}
\begin{tabular}{lccccc}
\toprule
Dataset Size & MMB & MathVista & MM-Vet & MMMU & Avg \\
\midrule
\rowcolor{colorful}
9M & \textbf{89.5} & \textbf{74.9} & \textbf{73.2} & \textbf{68.1} & \textbf{76.4} \\
7M & 89.4 & 74.7 & 73.0 & 67.8 & 76.2 \\
5M & 89.3 & 74.5 & 72.8 & 67.6 & 76.0 \\
3M & 88.9 & 73.6 & 71.9 & 66.6 & 75.3 \\
1M & 86.6 & 68.6 & 67.0 & 60.9 & 70.8 \\
0M & 84.6 & 64.4 & 62.8 & 56.0 & 67.0 \\
\bottomrule
\end{tabular}
}

\caption*{\scriptsize (g) Different LLMs}
\vspace{-3mm}
\centering
\resizebox{\linewidth}{!}{
\renewcommand{\tabcolsep}{3mm}
\begin{tabular}{lcccc}
\toprule
LLMs & MMLU & GPQA & MATH & Avg \\
\midrule
Qwen2-7B~\cite{yang2024qwen2} & 70.8 & 38.4 & 48.6 & 52.6 \\
\rowcolor{colorful}
Qwen2-7B-GenRecal (Llama-3-70B) & 78.4 & 38.9 & 49.1 & 55.5 \\
\cdashline{1-5}\noalign{\vskip 0.5ex}
Llama-3-70B~\cite{dubey2024llama} & \textbf{82.5} &\textbf{39.5} & \textbf{50.4} & \textbf{57.5} \\
\midrule
InternLM2.5-7B~\cite{cai2024internlm2} & 72.8 & 38.4 & 60.1 & 57.1 \\
\rowcolor{colorful}
InternLM2.5-7B-GenRecal (Qwen2.5-72B) & 83.7 & 43.6 & 61.8 & 63.0 \\
\cdashline{1-5}\noalign{\vskip 0.5ex}
Qwen2.5-72B~\cite{yang2024qwen2} & \textbf{86.1} & \textbf{45.9} & \textbf{62.1} & \textbf{64.7} \\
\bottomrule
\end{tabular}
}
\end{minipage}
\begin{minipage}[t]{0.32\linewidth}
\caption*{\scriptsize (b) Recalibrator Depth}
\vspace{-3mm}
\centering
\resizebox{\linewidth}{!}{
\renewcommand{\tabcolsep}{7.8mm}
\begin{tabular}{cccccc}
\hline
Depths & MMB & MathVista & MM-Vet & MMMU & Avg \\
\hline
20 & 89.9 & 75.1 & 73.6 & 68.3 & 76.7 \\
\cdashline{1-6}\noalign{\vskip 0.5ex}
15 & 89.8 & 75.2 & 73.5 & 68.4 & 76.7 \\
\cdashline{1-6}\noalign{\vskip 0.5ex}
10 & 89.6 & 75.0 & 73.4 & 68.3 & 76.6 \\
\cdashline{1-6}\noalign{\vskip 0.5ex}
5  & 89.4 & 74.8 & 73.3 & 68.2 & 76.4 \\
\cdashline{1-6}\noalign{\vskip 0.5ex}
\rowcolor{colorful}
2  & 89.5 & 74.9 & 73.2 & 68.1 & 76.4 \\
\cdashline{1-6}\noalign{\vskip 0.5ex}
1  & 88.0 & 73.0 & 72.0 & 66.5 & 74.9 \\
\hline
\end{tabular}
}

\caption*{\scriptsize (e) Fine-Tuning}
\vspace{-3mm}
\centering
\resizebox{\linewidth}{!}{
\renewcommand{\tabcolsep}{1mm}
\begin{tabular}{lcccccc}
\toprule
VLM & Finetuned & MMB & MathVista & MM-Vet & MMMU & Avg \\
\midrule
InternVL2.5-78B~\cite{chen2024expanding} & \xmark & 88.3 & 72.3 & 72.3 & 70.1 & 75.8 \\
InternVL2.5-78B-SFT & \cmark & \textbf{89.4} & \textbf{75.2} & \textbf{74.2} & \textbf{71.7} & \textbf{77.6} \\
\midrule
InternVL2.5-8B~\cite{chen2024expanding} & \xmark & 84.6 & 64.4 & 62.8 & 56.0 & 67.0 \\
InternVL2.5-8B-SFT & \cmark & 86.5 & 65.2 & 65.0 & 59.9 & 69.2 \\
\rowcolor{colorful}
InternVL2.5-8B-GenRecal & \cmark & \textbf{89.5} & \textbf{74.9} & \textbf{73.2} & \textbf{68.1} & \textbf{76.4} \\
\bottomrule
\end{tabular}
}

\caption*{\scriptsize (h) Different VLM Family}
\vspace{-3mm}
\centering
\resizebox{\linewidth}{!}{
\renewcommand{\tabcolsep}{0.5mm}
\begin{tabular}{lcccc}
\toprule
VLM & MM-Vet & MathVista & MMMU & Avg \\
\midrule
Ovis1.6-Gemma-9B~\cite{lu2024ovis} & 65.0 & 67.3 & 55.0 & 62.4\\
\rowcolor{colorful}
Ovis1.6-Gemma-9B-GenRecal (InternVL2.5-78B) & \textbf{74.0} & \textbf{75.3} & 67.7 & \textbf{72.3}\\
InternVL2.5-78B & 72.3 & 72.3 & \textbf{70.1} & 71.6 \\
\midrule
InternVL2.5-8B~\cite{chen2024expanding} & 62.8 & 64.4 & 56.0 & 61.1\\
\rowcolor{colorful}
InternVL2.5-8B-GenRecal (Ovis1.6-Gemma-27B) & \textbf{68.5} & \textbf{69.9} & 61.2 & \textbf{66.5}\\
Ovis1.6-Gemma-27B~\cite{lu2024ovis} & 67.2 & 69.7 & \textbf{61.7} & 66.2 \\
\bottomrule
\end{tabular}
}

\end{minipage}
\begin{minipage}[t]{0.32\linewidth}
\caption*{\scriptsize (c) Effect of NPE}
\vspace{-3mm}
\centering
\resizebox{\linewidth}{!}{
\renewcommand{\tabcolsep}{2mm}
\begin{tabular}{lcccccc}
\toprule
VLM & NPE & MMB & MathVista & MM-Vet & MMMU & Avg \\
\midrule
Qwen2-VL-7B-GenRecal & \xmark & 79.2 & 60.0 & 64.0 & 63.9 & 66.8 \\
\rowcolor{colorful}
Qwen2-VL-7B-GenRecal & \cmark & \textbf{88.4} & \textbf{68.8} & \textbf{70.4} & \textbf{65.6} & \textbf{73.3} \\
\midrule
InternVL2.5-8B-GenRecal & \xmark & 85.2 & 65.1 & 67.3 & 64.4 & 70.5 \\
\rowcolor{colorful}
InternVL2.5-8B-GenRecal & \cmark & \textbf{89.5} & \textbf{74.9} & \textbf{73.2} & \textbf{68.1} & \textbf{76.4} \\
\bottomrule
\end{tabular}
}

\caption*{\scriptsize (f) Cross-Tokenizer Methods}
\vspace{-3mm}
\centering
\resizebox{\linewidth}{!}{
\renewcommand{\tabcolsep}{0.5mm}
\begin{tabular}{lccccc}
\toprule
VLM & MathVista & MMB & MM-Vet & MMMU & Avg \\
\midrule
InternVL2.5-8B & 64.4 & 84.6 & 62.8 & 56.0 & 67.0 \\
InternVL2.5-8B-ULD~\cite{boizard2025towards} & 66.5 & 86.2 & 64.1 & 60.8 & 69.4 \\
InternVL2.5-8B-MOT~\cite{cui2025multi} & 68.4 & 87.5 & 65.0 & 62.3 & 70.8 \\
\rowcolor{colorful}
InternVL2.5-8B-GenRecal & \textbf{74.9} & \textbf{89.5} & \textbf{73.2} & \textbf{68.1} & \textbf{76.4} \\
\bottomrule
\end{tabular}
}

\caption*{\scriptsize (i) Recent VLMs}
\vspace{-3mm}
\centering
\resizebox{\linewidth}{!}{
\renewcommand{\tabcolsep}{5mm}
\begin{tabular}{lccccc}
\toprule
VLM & MMB & MathVista & MM-Vet & MMMU & Avg \\
\midrule
Qwen3-VL-8B~\cite{bai2025qwen3} & 86.8 & 77.2 & 74.5 & 69.6 & 77.0 \\
\rowcolor{colorful}
Qwen3-VL-8B (Qwen3-VL-32B) & \textbf{89.9} & 80.1 & \textbf{77.6} & \textbf{72.7} & \textbf{80.1} \\
\rowcolor{colorful}
Qwen3-VL-8B (InternVL3.5-38B) & 89.7 & \textbf{80.3} & 77.4 & 72.5 & 80.0 \\
\midrule
InternVL3.5-8B~\cite{wang2025internvl3} & 86.5 & 78.4 & 83.1 & 73.4 & 80.4 \\
\rowcolor{colorful}
InternVL3.5-8B (Qwen3-VL-32B) & \textbf{89.6} & \textbf{81.5} & 86.0 & 76.3 & \textbf{83.4} \\
\rowcolor{colorful}
InternVL3.5-8B (InternVL3.5-38B) & 89.4 & 81.3 & \textbf{86.2} & \textbf{76.5} & \textbf{83.4} \\
\midrule
Qwen3-VL-32B~\cite{bai2025qwen3} & 90.6 & 83.8 & 79.4 & 76.0 & 82.5 \\
InternVL3.5-38B~\cite{wang2025internvl3} & 90.3 & 81.9 & 82.2 & 76.9 & 82.8 \\
\bottomrule
\end{tabular}
}

\end{minipage}
\vspace{-5mm}
\end{table}

\subsection{Discussion}

\noindent\textbf{Representation Mismatch Beyond Tokenizer Mismatch.}
Even within a family the tokenizer can differ---InternVL2.5-8B uses InternLM2.5~\cite{cai2024internlm2} while InternVL2.5-78B uses Qwen2.5~\cite{yang2024qwen2}, whereas Qwen2-VL-7B/72B~\cite{wang2024qwen2vl} share the Qwen2 tokenizer (\cref{fig:1}). More importantly, even with an \emph{identical} tokenizer, a large teacher--student gap causes a \emph{representation} mismatch: conventional distillation routes knowledge through the low-dimensional student head, creating a bottleneck. GenRecal instead distills through the high-capacity teacher head, with the Recalibrator warm-starting student features into the teacher space (\cref{fig:3}). This is why it helps even under the same tokenizer: in \cref{tab:4} (Qwen2-VL-72B teacher, 7B/2B students), GenRecal beats both SFT and traditional KD---MiniLLM~\cite{gu2024minillm} (reverse KL), DistiLLM~\cite{ko2024distillm} (skewed KL), and LLaVA-KD~\cite{cai2024llava} (3-stage KL)---by a clear margin, showing explicit feature alignment matters even with shared token types.

\begin{table}[t!]
\centering
\vspace{-3mm}
\caption{Comparison of computational cost (FLOPs and Training Time) and performance across cross-tokenizer distillation methods. We average performance across four benchmarks: MMB~\cite{liu2023mmbench}, MathVista~\cite{lu2023mathvista}, MM-Vet~\cite{yu2023mm}, and MMMU~\cite{yue2023mmmu}.}
\label{tab:compute}
\vspace{-3mm}
\resizebox{\linewidth}{!}{
\renewcommand{\tabcolsep}{1mm}
\begin{tabular}{llccrc}
\toprule
Method &  Extra & FLOPs & Training Stages & Training Time & Avg. Perform.\\
\midrule

GenRecal 
& Recalibrator
& 298.3$\times10^{12}$ 
& 3
& 3.1+10.4+5.3=\textbf{18.8} hours
& \textbf{76.4}\\

ULD~\cite{boizard2025towards} 
& Vocab Matching 
& 296.1$\times10^{12}$ 
& 1
& 11.1$\times$2(Epochs)=22.2 hours
& 69.4\\

MOT~\cite{cui2025multi} 
& Optimal Transport 
& 296.1$\times10^{12}$ 
& 1
& 12.5$\times$2(Epochs)=25.0 hours 
& 70.8\\
\bottomrule
\end{tabular}
}
\vspace{-3mm}
\end{table}


\vspace{1.5mm}
\noindent\textbf{Cross-Token Matching.}
Cross-token matching takes two forms: (a) word-embedding matching, infeasible without retraining the decoder, and (b) semantic feature matching. For (b), ULD~\cite{boizard2025towards} and MOT~\cite{cui2025multi} align logit-space probability mass via Wasserstein distance or optimal transport (\cref{tab:5}(f)), but cannot handle token-split indices and resort to zero-padding or truncation, losing information. GenRecal avoids logit-level matching altogether: the Recalibrator aligns features \emph{before} the head so the teacher \textit{VLM-head} reads student features directly, needing no token correspondence, probability sorting, padding, or transport cost. It adds only a small FLOPs fraction (\cref{tab:5}(a)), so GenRecal's total cost is comparable to ULD/MOT yet reaches higher accuracy (\cref{tab:compute}; same data, InternVL2.5-8B student and -78B teacher).

\vspace{1.5mm}
\noindent\textbf{Why a Non-Linear Recalibrator, and Where the Novelty Lies.}
Prior shared-space results rely on \emph{linear} word-embedding maps~\cite{mikolov2013exploiting, smith2017offline, conneau2017word}, which act on input embeddings, not on the answer-conditioned hidden states after the \textit{VLM-body}. Aligning these post-body features while preserving token order exceeds a linear map, so we use shallow transformer decoder blocks. The novelty is thus not the module itself---depth beyond two layers saturates (\cref{tab:5}(b))---but the \emph{stable recipe} that makes cross-tokenizer teacher-head distillation work.

\begin{table}[t!]
\centering
\caption{Decomposing the source of GenRecal's improvement on Qwen2-VL-7B with the same-token-type teacher Qwen2-VL-72B, so that traditional logit-KD baselines remain valid. All rows use the same 9M data. Avg.\ is over MMB~\cite{liu2023mmbench}, MathVista~\cite{lu2023mathvista}, MM-Vet~\cite{yu2023mm}, and MMMU~\cite{yue2023mmmu}.}
\label{tab:decomp}
\vspace{-3mm}
\resizebox{\linewidth}{!}{
\renewcommand{\tabcolsep}{8mm}
\begin{tabular}{lcc}
\toprule
Setup (all use the \emph{same} 9M data) & Avg.\ & $\Delta$ \\
\midrule
Baseline Qwen2-VL-7B (no training)                              & 64.3 & --     \\
\,$+$ 9M SFT (Stage-3 recipe only, no distill, no Recalib.)     & 66.3 & $+2.0$ \\
\,$+$ multi-stage SFT$+$logit KD (LLaVA-KD recipe, no Recalib.) & 67.7 & $+3.4$ \\
\rowcolor{colorful}
\,$+$ GenRecal (Recalibrator)                             & \textbf{72.3} & \textbf{$+8.0$} \\
\bottomrule
\end{tabular}
}
\vspace{-3mm}
\end{table}

\begin{table}[t!]
\centering
\caption{Stage-wise checkpoints of GenRecal with InternVL2.5-8B (student) and InternVL2.5-78B (teacher), compared against an SFT-only baseline on the same data. Stage~1 trains only the Recalibrator and leaves the student unchanged.}
\label{tab:stage}
\vspace{-3mm}
\resizebox{\linewidth}{!}{
\renewcommand{\tabcolsep}{1mm}
\begin{tabular}{llcccccc}
\toprule
Checkpoint & Trainable parameters & Data & MMB & MathVista & MM-Vet & MMMU & Avg  \\
\midrule
Baseline                        & ---                       & ---  & 84.6 & 64.4 & 62.8 & 56.0 & 67.0 \\
After Stage~1                   & Recalibrator only (align) & $9$M & 84.6 & 64.4 & 62.8 & 56.0 & 67.0 \\
After Stage~2                   & Recalibrator $+$ Student  & $9$M & 88.2 & 72.5 & 70.9 & 65.7 & 74.3 \\
\rowcolor{colorful}
After Stage~3 (GenRecal full)   & Student only (SFT polish) & $6$M & \textbf{89.5} & \textbf{74.9} & \textbf{73.2} & \textbf{68.1} & \textbf{76.4} \\
\midrule
SFT only                        & Student only (SFT polish) & $9$M & 86.5 & 65.2 & 65.0 & 59.9 & 69.2 \\
\bottomrule
\end{tabular}
}
\vspace{-3mm}
\end{table}

\begin{table}[t!]
\centering
\caption{Effect of training-data scale on GenRecal with InternVL2.5-8B (student) and InternVL2.5-78B (teacher), compared against SFT-only at full scale. Stage~3 uses a random $\sim$70\% subsample of the Stage~2 data.}
\label{tab:scale}
\vspace{-3mm}
\resizebox{\linewidth}{!}{
\renewcommand{\tabcolsep}{2mm}
\begin{tabular}{lcccccccc}
\toprule
Setting & Align data & Distill data & Final SFT data & MMB & MathVista & MM-Vet & MMMU & Avg \\
\midrule
Baseline (no train)     & --- & ---  & ---    & 84.6 & 64.4 & 62.8 & 56.0 & 67.0 \\
SFT only                & --- & ---  & $9$M   & 86.5 & 65.2 & 65.0 & 59.9 & 69.2 \\
\midrule
GenRecal @ $1$M         & $1$M & $1$M & $0.7$M & 86.6 & 68.6 & 67.0 & 60.9 & 70.8 \\
GenRecal @ $3$M         & $3$M & $3$M & $2$M   & 88.9 & 73.6 & 71.9 & 66.6 & 75.3 \\
GenRecal @ $5$M         & $5$M & $5$M & $3$M   & 89.3 & 74.5 & 72.8 & 67.6 & 76.0 \\
\rowcolor{colorful}
GenRecal @ $9$M (full)  & $9$M & $9$M & $6$M   & \textbf{89.5} & \textbf{74.9} & \textbf{73.2} & \textbf{68.1} & \textbf{76.4} \\
\bottomrule
\end{tabular}
}
\vspace{-3mm}
\end{table}


\vspace{1.5mm}
\noindent\textbf{Isolating the Recalibrator from the Recipe.}
To check whether the gains come from the Recalibrator or merely from the multi-stage recipe and final SFT, we disentangle them on Qwen2-VL-7B with the same-token-type teacher Qwen2-VL-72B (so logit-KD baselines stay valid), all on the same 9M data. The recipe alone (SFT$+$logit-KD, no Recalibrator) gives $+3.4$ Avg, while adding the Recalibrator adds a further $+4.6$ (\cref{tab:decomp}). Nor is the gain from Stage-3 SFT: after Stage~2, \emph{before any final SFT}, GenRecal already reaches $74.3$ Avg vs.\ $69.2$ for SFT-only on the same 9M (\cref{tab:stage}). Nor is it mere data scale: GenRecal@1M ($\sim$11\% data) already beats SFT@9M, and @3M nears the full result (\cref{tab:scale}). Further analyses are provided in the appendix: open-weight teacher dependence and the SFT fallback (\appref{app:F}), component-level ablations (\appref{app:G}), the regularization formulation (\appref{app:H}), error-profile and calibration (\appref{app:I}), and a compute/fairness comparison against baselines (\appref{app:J}). Note that GenRecal operates under \emph{open-weight} teacher access, where the teacher's parameters and hidden features are available for training.
\section{Conclusion}

We present \textbf{Gen}eration after \textbf{Recal}ibration (GenRecal), a general-purpose distillation framework that transfers knowledge from large to small vision-language models (VLMs) even when they adopt different token types. Existing distillation methods assume that the teacher and student share the same token types, while GenRecal removes this constraint through a Recalibrator that aligns the student's hidden features into the teacher's representation space. Since the Recalibrator is discarded after training, inference preserves the student's original architecture and cost. Our analyses further indicate that these gains stem primarily from the recalibration mechanism rather than from the multi-stage recipe, and that feature alignment remains beneficial even when the teacher and student already share a tokenizer. For future work, we plan to extend the Recalibrator to intermediate layers to capture fine-grained, sequential knowledge, and to explore distillation from multiple VLM sources. We envision GenRecal as a pivotal and versatile framework for knowledge distillation.

\section*{Acknowledgements}
We thank our colleagues at NVIDIA, in particular Sharath Turuvekere Sreenivas, for valuable discussions on cross-tokenizer and cross-model knowledge distillation. We further thank Sepehr Sameni and Daniel Korzekwa for experimenting with this approach on Minitron and Cosmos families; subsequently, Sharath Turuvekere Sreenivas and Pavlo Molchanov developed the follow-up X-Token~\cite{sreenivas2026x}, extending it to multi-teacher cross-tokenizer distillation on the logit space.

%
%
\bibliographystyle{splncs04}
\bibliography{main}

\clearpage
\onecolumn

\appendix

\section{Token Type Examples}
\phantomsection\label{app:A}
\begin{figure*}[h!]
    \centering
    \includegraphics[width=0.9\textwidth]{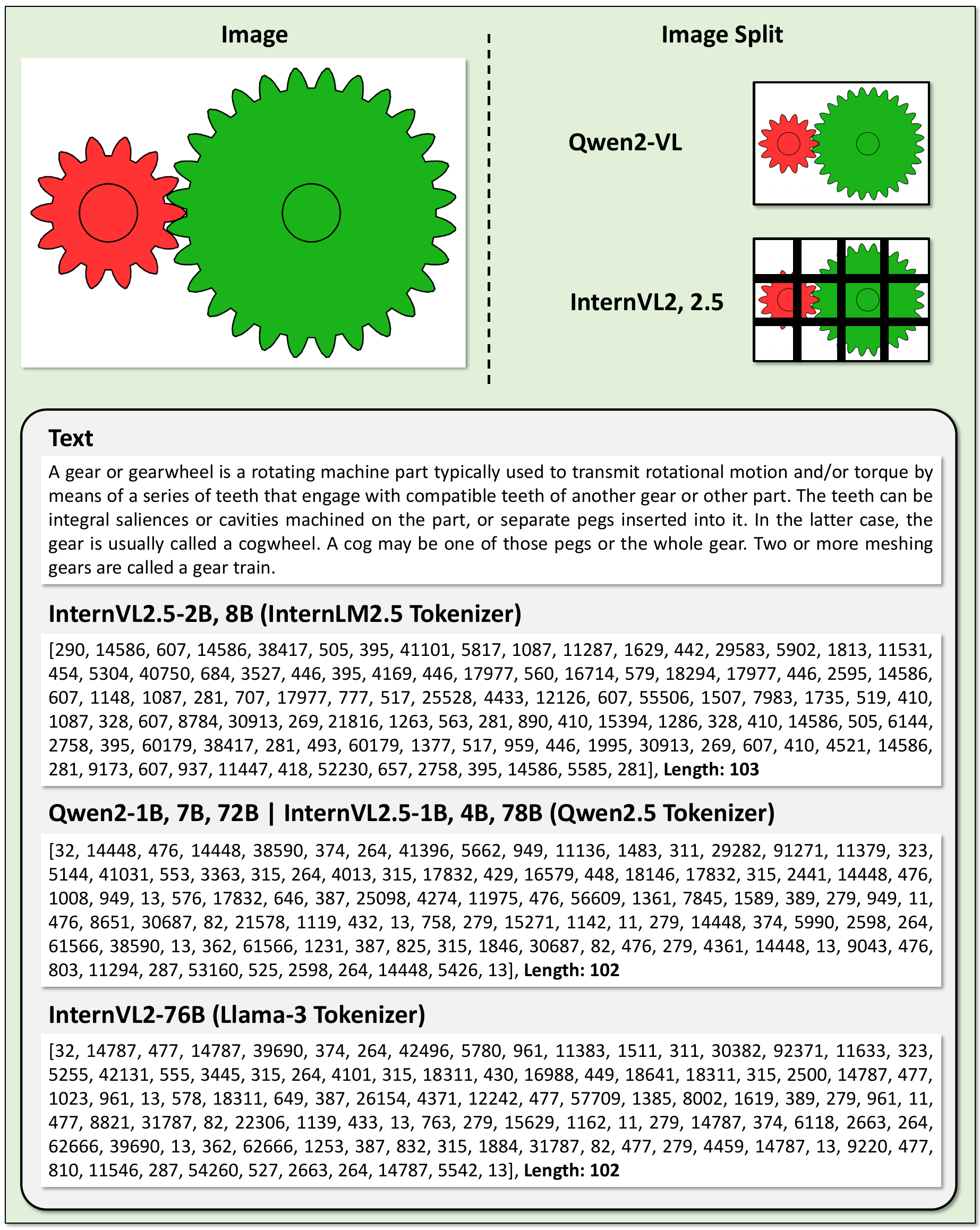}
    \label{fig:AppA}
\end{figure*}

\clearpage
\section{Possible Distillation Pair for VLM Combinations: Traditional Distillation versus GenRecal}
\phantomsection\label{app:B}
\begin{figure*}[h!]
    \centering
    \includegraphics[width=\textwidth]{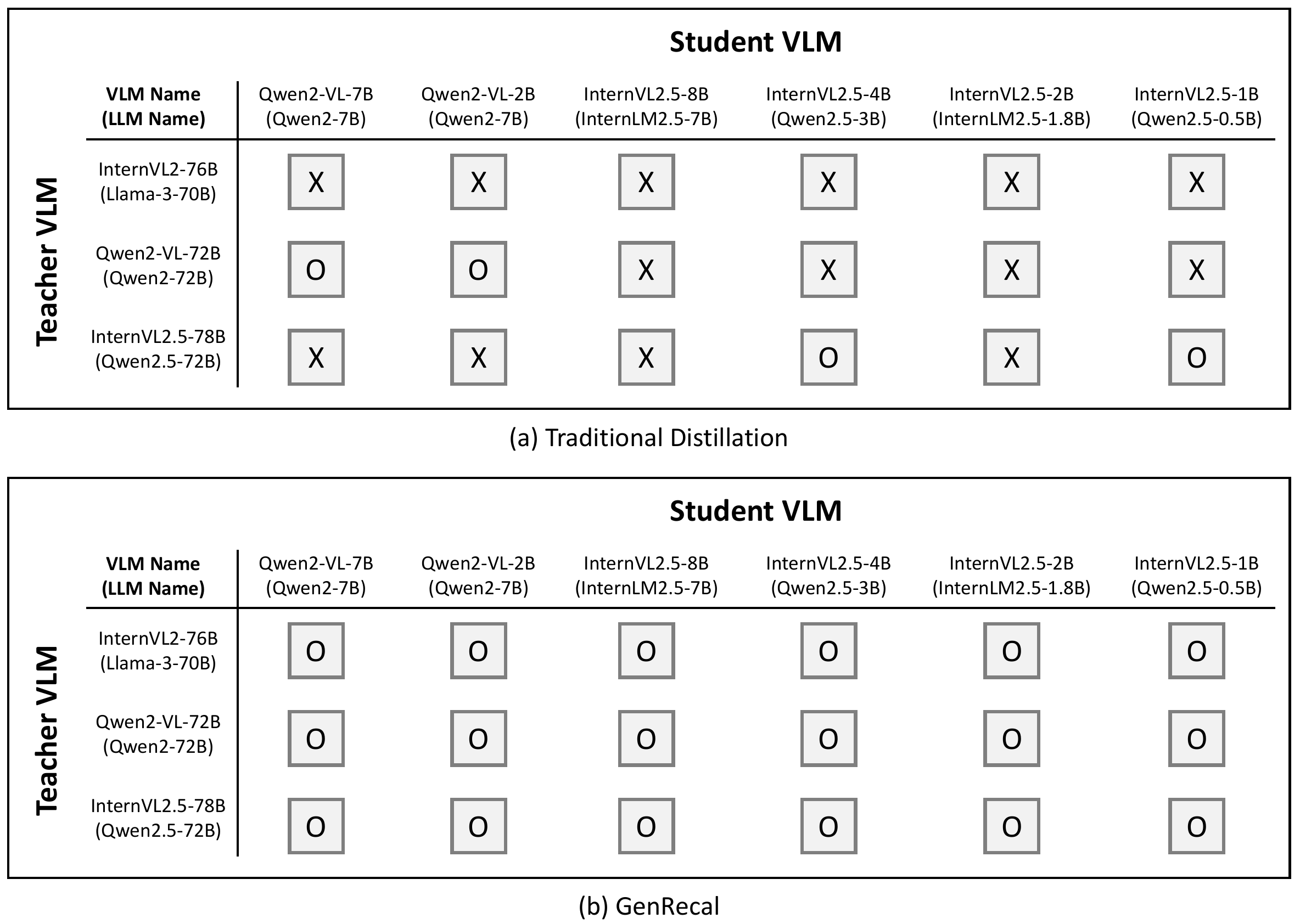}
    \caption{We explore the range of distillation combinations between teacher and student VLMs using two approaches: (a) traditional distillation~\cite{cai2024llava} and (b) our proposed model, GenRecal. Unlike traditional distillation—which supports only a limited set of pairings—GenRecal offers the flexibility to select any model for distillation, thereby enabling a more versatile and comprehensive distillation framework.}
    \label{fig:AppB}
\end{figure*}

\clearpage
\section{Extended Related Works}
\phantomsection\label{app:C}
\noindent{\textbf{Evolution of Vision-Language Models.}} Since the emergence of visual instruction tuning~\cite{liu2023visual}, many variations of VLMs have emerged, such as LLaVA-1.5~\cite{liu2023improved}, InstructBLIP~\cite{dai2023instructblip}, ShareGPT-4V~\cite{chen2023sharegpt4v}, and MiniGPT-v2~\cite{chen2023minigpt, zhu2023minigpt}, typically starting at a standard model size of 7B. After that period, numerous VLMs~\cite{liu2024llavanext, mckinzie2024mm1, li2024mini, dong2024internlm4khd} have divided an image into multiple sub-regions to focus on the image's details so that visual perception is enhanced. CoLLaVO~\cite{lee-etal-2024-collavo} and MoAI~\cite{10.1007/978-3-031-72967-6_16} utilize computer vision models directly for visual capability, and Mini-Gemini~\cite{li2024mini}, MoVA~\cite{zong2024mova}, and Eagle~\cite{shi2024eagle} employ multiple vision encoders such as CLIP~\cite{clip}, ConvNext~\cite{woo2023convnext}, DINO-v2~\cite{oquab2023dinov2}, and SAM~\cite{kirillov2023segment}. In parallel, Meteor~\cite{NEURIPS2024_473a9a75} explores the efficient way of learning complex reasoning abilities, and TroL~\cite{lee-etal-2024-trol} and Phantom~\cite{lee2024phantom} investigate propagation modification for how we can embed vision-language knowledge as much as possible despite using the same architectures. More recently, many VLMs like Molmo-72B~\cite{deitke2024molmo}, LLaVA-OneVision-72B~\cite{li2024llava}, NVLM-72B~\cite{nvlm2024}, Qwen2-VL-72B~\cite{wang2024qwen2vl} and InternVL2.5-78B~\cite{chen2024expanding} have employed large-scale language models such as Qwen2/Qwen2.5-72B~\cite{yang2024qwen2}. Thanks to scaling laws~\cite{kaplan2020scaling, chung2022scaling}, they have closely reached the performances of GPT-4V~\cite{gptsyscard} and Claude-3.5 Sonnet~\cite{claude3series2024}. Therefore, developing VLMs with large-scale language models is getting to a standard these days. Beyond architectural design, recent efforts further explore training and evaluating such VLMs, including unified reinforcement and imitation learning~\cite{NEURIPS2025_e5849736}, recursive think--answer reasoning~\cite{Lee_2026_CVPR_Recursive}, multimodal agentic reasoning~\cite{kang2026agent, cho2026spatialclawrethinkingactioninterface}, and multi-turn conversation benchmarking~\cite{Lee_2025_ICCV}.

\vspace{1.5mm}
\noindent{\textbf{Knowledge Distillation.}} This provides an effective framework~\cite{hinton2015distilling} for transferring the rich representations of large, high-performing models to smaller, efficient counterparts while preserving performance.  Early studies primarily focused on aligning the output logits of a ``teacher'' model with those of a ``student''~\cite{sanh2019distilbert, turc2019well}. Subsequently, the FitNet framework~\cite{romero2014fitnets} extended this idea by supervising the student using intermediate feature representations rather than final outputs. 
In this setup, convolutional layers project the student’s features to resemble those of the teacher, and discrepancies are minimized using an L2 loss. 

Following this feature-based paradigm, later methods explored more sophisticated ways of leveraging internal representations for effective knowledge transfer. 
For instance, \cite{sun2019patient, chen2021distilling, wang2024crosskd, ben2022s} demonstrated that incorporating multiple intermediate layers from the teacher significantly enhances distillation effectiveness. 
Probabilistic approaches such as PKT~\cite{passalis2020probabilistic} reformulated the teacher’s knowledge as a probability distribution, aligning it with the student’s outputs via KL divergence. 
Similarly, RKD~\cite{park2019relational} captured inter-sample relationships, while CRD~\cite{tian2019contrastive} combined contrastive learning with traditional distillation. 

Multi-stage and relational strategies further broadened the field. 
AT~\cite{zagoruyko2016paying} utilized attention maps from several teacher layers, and FSP~\cite{yim2017gift} employed matrices derived from feature maps to guide the student. 
Later, SP~\cite{tung2019similarity} evaluated similarity among input samples, and OFD~\cite{heo2019comprehensive} introduced a margin-based distance metric derived from ReLU activations to capture essential information. 

As the methodology matured, two main paradigms emerged. 
\textit{On-policy} approaches~\cite{austin2021program, ko2024distillm, agarwal2024policy, gu2024minillm} dynamically sample data during training, whereas \textit{off-policy} methods~\cite{ko2024distillm, sun2019patient, turc2019well, zhang2024dual, boizard2025towards} rely on pre-collected datasets. 
While early research mostly adopted single-teacher frameworks~\cite{agarwal2024policy, gu2024minillm, shu2024llava, muralidharan2024compact}, multi-teacher distillation has gained increasing attention despite challenges such as architectural differences, vocabulary mismatches, and task misalignment~\cite{timiryasov2023baby, lee2023ensemble, wan2024fusechat}. 
Representative examples include DKMF~\cite{wang2021distilling} and FNKD~\cite{xu2020feature}, which utilize teacher feature representations, while DGKD~\cite{son2021densely} aggregates knowledge from multiple teachers to boost student performance. 

Recent work has also emphasized refining loss functions used in distillation. 
MiniLLM~\cite{gu2024minillm} and DistiLLM~\cite{ko2024distillm} propose modified KL divergences---such as reverse or skewed variants---to reduce overfitting, particularly for long-tail predictions. 
Similarly, \cite{wu2024rethinking} introduced a dynamic loss-balancing scheme that adaptively combines conventional and reverse KL divergences during training. 
Finally, to enhance reasoning capabilities, recent studies integrate chain-of-thought (CoT) supervision. 
Methods such as~\cite{hsieh2023distilling, tian2024beyond} leverage detailed reasoning traces from larger models, while TinyLLM~\cite{tian2024beyond} aggregates multi-teacher reasoning to enrich training signals and improve generalization.

\vspace{1.5mm}
\noindent{\textbf{Efficient AI for Vision-Language Models.}} In the vision-language setting, recent works distill or compress large VLMs into smaller, deployable ones: VLsI~\cite{Lee_2025_CVPR} transfers knowledge through verbalized layer-to-layer interactions, masking-teacher with student reinforcement~\cite{Lee_2026_CVPR_Masters} and reasoning-prefix masking~\cite{yu2026hide} refine the teacher--student signal, visual-token pruning~\cite{kim2026and} removes redundant visual computation, and broader recipes merge, modify, and distill VLMs~\cite{lee2025building}. For language models, related efforts study prompt-based teaching~\cite{lee2026zone}, skill-of-mind conversational modeling~\cite{leeenhancing}, and refinement-capability evaluation~\cite{lee2026refinebench}. The authors' broader research additionally spans robustness~\cite{lee2021towards, NEURIPS2021_8e5e15c4, Lee_2022_CVPR, Lee_2023_ICCV, lee2020training} and causal representation learning~\cite{Kim_2023_CVPR, KIM2026112173, 10222502}.

\clearpage
\section{Loss Graphs for Recalibrator}
\phantomsection\label{app:D}
\begin{figure*}[h!]
    \centering
    \includegraphics[width=\textwidth]{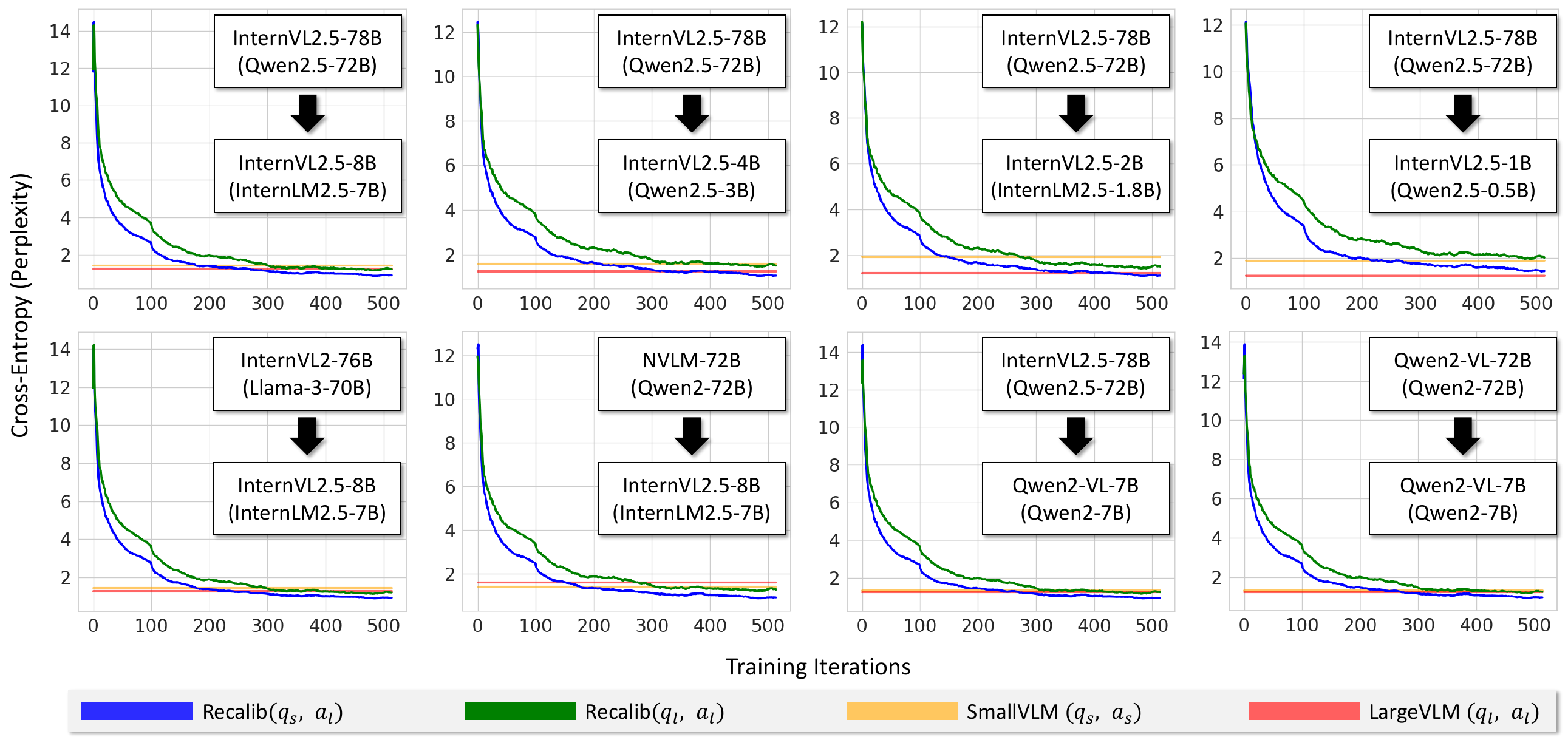}
    \vspace{-5mm}
    \caption{Illustrating the loss graphs of training Recalibrator, where we deal with various combinations of VLMs: NVLM~\cite{nvlm2024}, Qwen2-VL~\cite{wang2024qwen2vl}, InternVL2~\cite{chen2023internvl}, and InternVL2.5~\cite{chen2024expanding}. Note that, the parenthesis in the figure means the name of LLM used in VLMs. `Recalib$(q_s,a_l)$' and `Recalib$(q_l,a_l)$' represent the cross entropy loss with Recalibrator logits and \textit{VLM-head} of large VLM (see Section 3.2). `SmallVLM$(q_s,a_s)$' and `LargeVLM$(q_l,a_l)$' means original cross entropy loss for SFT without Recalibrator. They are not used in training Recalibrator. They just represent the averaged cross entropy (perplexity) during the whole training to compare them with `Recalib$(q_s,a_l)$' and `Recalib$(q_l,a_l)$'.}
    \label{fig:AppD}
\end{figure*}

\clearpage
\section{Dataset Composition and Analysis}
\phantomsection\label{app:E}
We gather a 9M visual instruction tuning dataset covering a wide range of vision-language capabilities, such as general visual question answering, dense image captioning, chart/diagram/document understanding, commonsense knowledge, science and math understanding, and multi-dimensional reasoning. Our dataset includes LLaVA-OneVision~\cite{li2024llava}, MMC~\cite{liu2023mmc}, DenseFusion~\cite{li2024densefusion}, Cambrian~\cite{tong2024cambrian}, GPT-4V-filtered synthetic data of SA-1B~\cite{kirillov2023segment} and Infinity-MM~\cite{gu2024infinity}, Finance-QA~\cite{Sujet-Finance-QA-Vision-100k}, Wikipedia knowledge~\cite{li2024cvlm}, InfoSeek~\cite{chen2023can}, science \& mathematical reasoning (SMR)~\cite{zhang2024beyond}, document-downstream/reasoning~\cite{hu2024mplug}, WildVision~\cite{lu2024wildvision}, SROIE~\cite{huang2019icdar2019}, RLAI-F~\cite{yu2024rlaif}, M3CoT~\cite{chen-etal-2024-m3cot}, LLaVAR~\cite{zhang2023llavar}, KonIQ~\cite{hosu2020koniq}, iNaturalist2018~\cite{van2018inaturalist}. For further analysis, we particularly categorize the visual instruction tuning dataset that we use to build GenRecal into three domains: `Knowledge', `Science \& Math', and `Chart \& Document'. Based on these categories, we conduct dataset composition analysis by removing each of the three domains. This implies that MMMU~\cite{yue2023mmmu} requires VLMs' ability for the domain of `Knowledge' more than the others, while MathVista~\cite{lu2023mathvista} requires their capabilities for `Science \& Math'.

\vspace{-5mm}
\begin{figure}[h!]
    \centering
    \includegraphics[width=\textwidth]{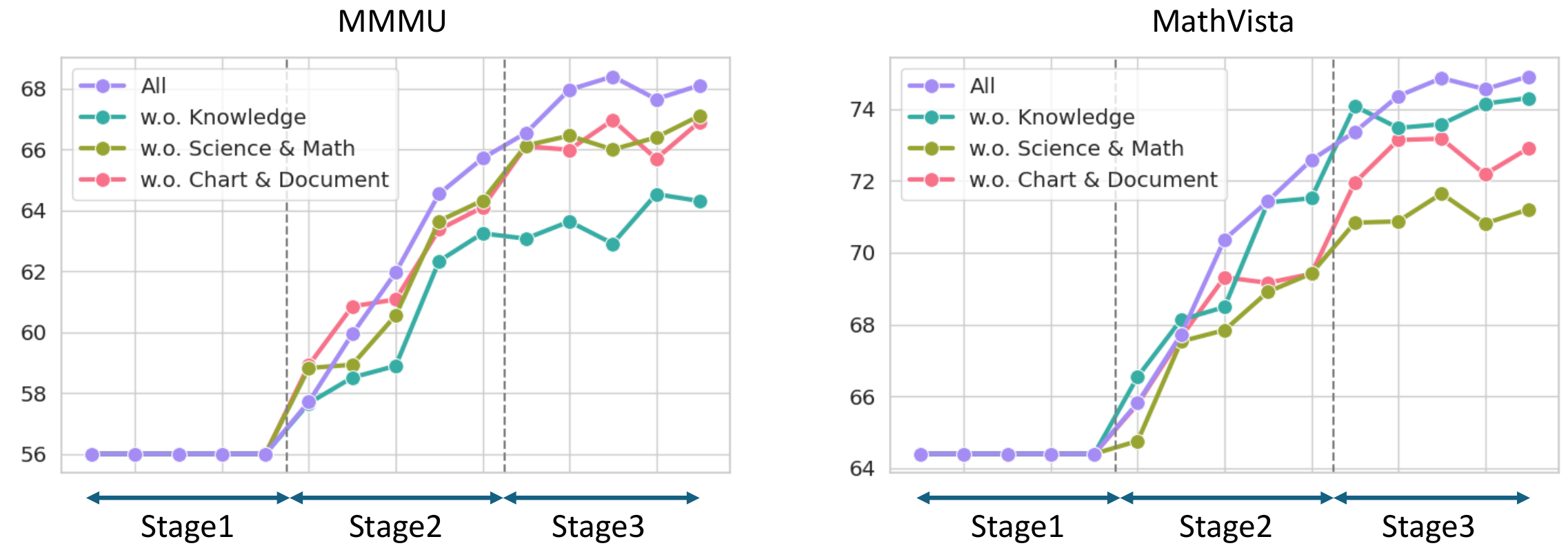}
    \caption{Accuracy trends across different datasets: (Left) MMMU~\cite{yue2023mmmu}, and (Right) MathVista~\cite{lu2023mathvista} over three training stages. The performance is shown for different category of visual instruction tuning dataset: ``All'' (without any exclusion), ``w.o. Knowledge'', ``w.o. Science \& Math'', and ``w.o. Chart \& Document''. The results indicate the impact of these exclusions on accuracy progression, highlighting how the absence of specific knowledge domains affects. Note that, Note that each data point within the same stage represents 20\% of the training progress, with five data points measured per stage.}
    \label{fig:AppE}
\end{figure}
\vspace{-5mm}

\begin{itemize}
    \item \textbf{Knowledge}: GPT-4V filtered synthetic of SA-1B~\cite{kirillov2023segment} and Infinity-MM~\cite{gu2024infinity}, Wikipedia Knowledge~\cite{li2024cvlm}, Wild-Vision, InfoSeek~\cite{chen2023can}, WildVision~\cite{lu2024wildvision}, and iNaturalist2018~\cite{van2018inaturalist}
    \item \textbf{Science \& Math}: LLaVA-OneVision (subset)~\cite{li2024llava}, Cambrian (subset)~\cite{tong2024cambrian}, Finance-QA~\cite{Sujet-Finance-QA-Vision-100k}, SMR~\cite{zhang2024beyond}, M3CoT~\cite{chen-etal-2024-m3cot}, and KonIQ~\cite{hosu2020koniq}
    \item \textbf{Chart \& Document}: LLaVA-OneVision (subset)~\cite{li2024llava}, Cambrian (subset)~\cite{tong2024cambrian}, MMC~\cite{liu2023mmc}, RLAI-F~\cite{yu2024rlaif}, document-downstream/reasoning~\cite{hu2024mplug}, SROIE~\cite{huang2019icdar2019}, and LLAVAR~\cite{zhang2023llavar}
\end{itemize}

\clearpage
\section{Open-Weight Teacher Dependence and SFT Fallback}
\phantomsection\label{app:F}
GenRecal trains with the teacher's hidden states and language head, so it assumes an \emph{open-weight} teacher. This matches our central motivation---distilling heterogeneous \emph{open} VLMs with different token types---rather than closed/API models. When only a closed/API teacher is available, the natural fallback is supervised fine-tuning (SFT) on teacher-generated responses. \cref{tab:sftfallback} compares this fallback with GenRecal on two students. Notably, a substantial portion of our 9M corpus already consists of closed/API-generated samples (\eg, Infinity-MM~\cite{gu2024infinity}, LLaVA-OneVision~\cite{li2024llava}, Cambrian~\cite{tong2024cambrian}, and RLAI-F~\cite{yu2024rlaif}; see Appendix~E). Thus the ``SFT only'' row is effectively trained on closed/API-generated responses, yet GenRecal still outperforms it by $+6.0$ and $+7.2$ Avg. We therefore scope the general-purpose claim to open-weight teachers and state the teacher-access assumption explicitly in the revision.

We additionally clarify our response-collection setup. Some 9M samples contain very short answers (\eg, ``A'' or ``32''). To provide a richer distillation signal, we collect correct large-teacher responses with more detailed reasoning before training, and all experiments are based on this setup.

\begin{table}[h!]
\centering
\caption{Open-weight teacher dependence and the SFT fallback for closed/API teachers. ``SFT only'' is trained on closed/API-generated responses drawn from the same 9M corpus, yet GenRecal still outperforms it by a large margin.}
\label{tab:sftfallback}
\vspace{-1mm}
\resizebox{\linewidth}{!}{
\renewcommand{\tabcolsep}{3mm}
\begin{tabular}{lcccccccccc}
\toprule
& \multicolumn{5}{c}{\textbf{Qwen2-VL-7B student}} & \multicolumn{5}{c}{\textbf{InternVL2.5-8B student}} \\
\cmidrule(lr){2-6}\cmidrule(lr){7-11}
Method & MMB & MathVista & MM-Vet & MMMU & Avg & MMB & MathVista & MM-Vet & MMMU & Avg \\
\midrule
Baseline    & 83.0 & 58.2 & 62.0 & 54.1 & 64.3 & 84.6 & 64.4 & 62.8 & 56.0 & 67.0 \\
w/ SFT only & 84.3 & 60.5 & 64.2 & 56.3 & 66.3 & 86.5 & 65.2 & 65.0 & 59.9 & 69.2 \\
\rowcolor{colorful}
w/ GenRecal & \textbf{87.8} & \textbf{69.5} & \textbf{67.8} & \textbf{64.2} & \textbf{72.3} & \textbf{89.5} & \textbf{74.9} & \textbf{73.2} & \textbf{68.1} & \textbf{76.4} \\
\midrule
$\Delta$ (GenRecal$-$SFT) & +3.5 & +9.0 & +3.6 & +7.9 & \textbf{+6.0} & +3.0 & +9.7 & +8.2 & +8.2 & \textbf{+7.2} \\
\bottomrule
\end{tabular}
}
\end{table}


\clearpage
\section{Component-Level Ablation of GenRecal}
\phantomsection\label{app:G}
We complement the stage-wise analysis (\cref{tab:stage}) with a component-level view in \cref{tab:component}, where each row removes a single component of GenRecal and the corresponding source ablation in the main text is cited. Removing the autoregressive loss $\mathcal{L}_{ar}$ is by far the most damaging, as it collapses feature alignment; removing the regularization term is the next most harmful. The new positional embedding (NPE) and a deeper \textit{Rec-body} provide smaller but consistent improvements. This confirms that the Recalibrator and its training objectives---not the multi-stage SFT recipe---are responsible for the bulk of the gain.

\begin{table}[h!]
\centering
\caption{Component-level ablation of GenRecal (InternVL2.5-8B student, InternVL2.5-78B teacher). Each row removes one component; the last column cites the source ablation in the main text. Avg.\ is over MMB / MathVista / MM-Vet / MMMU.}
\label{tab:component}
\vspace{-1mm}
\resizebox{\linewidth}{!}{
\renewcommand{\tabcolsep}{4mm}
\begin{tabular}{lcccccl}
\toprule
Variant & MMB & MathVista & MM-Vet & MMMU & Avg & Source \\
\midrule
Baseline (no teacher signal)                   & 84.6 & 64.4 & 62.8 & 56.0 & 67.0 & --- \\
w/ SFT only (no Recalibrator)                  & 86.5 & 65.2 & 65.0 & 59.9 & 69.2 & \cref{tab:5}(e) \\
\midrule
GenRecal w/o $\mathcal{L}_{ar}$ (KL only)      & 68.9 & 55.2 & 55.4 & 50.6 & 57.5 & \cref{tab:arloss} \\
GenRecal w/o regularization                    & 88.2 & 69.8 & 63.5 & 58.9 & 70.1 & \cref{tab:3} \\
GenRecal w/o NPE                               & 85.2 & 65.1 & 67.3 & 64.4 & 70.5 & \cref{tab:5}(c) \\
GenRecal w/ depth-1 \textit{Rec-body} (vs.\ 2) & 88.0 & 73.0 & 72.0 & 66.5 & 74.9 & \cref{tab:5}(b) \\
\rowcolor{colorful}
GenRecal (full)                                & \textbf{89.5} & \textbf{74.9} & \textbf{73.2} & \textbf{68.1} & \textbf{76.4} & --- \\
\bottomrule
\end{tabular}
}
\end{table}


\clearpage
\section{Regularization Formulation Ablation}
\phantomsection\label{app:H}
The regularization loss $\mathcal{L}_{reg}$ passes only the teacher's own tokens $[z_{q_l}, z_{a_l}]$ through the Recalibrator and constrains the result to remain readable by the teacher head (\cref{alg:2}). To motivate this specific formulation, \cref{tab:regform} ablates alternative constraints on the same teacher input. A pointwise feature constraint (Feature MSE) that simply keeps the Recalibrator near identity helps only marginally, because matching features in $\ell_2$ does not guarantee they remain interpretable by the language head. Enforcing the constraint \emph{through} the teacher head---either as a KL or a CE term---is substantially more effective, and combining both (our choice) is best. This explains why the teacher-only pass anchors the Recalibrator on the teacher manifold and prevents drift, complementing the cosine-diagonal analysis in \cref{fig:7}.

\begin{table}[h!]
\centering
\caption{Ablation over the form of the regularization loss $\mathcal{L}_{reg}$ applied to the teacher input $[z_{q_l}, z_{a_l}]$ (InternVL2.5-8B student, InternVL2.5-78B teacher). Constraining the Recalibrator output \emph{through} the teacher head is far more effective than a pointwise feature constraint. Avg.\ is over MMB / MathVista / MM-Vet / MMMU.}
\label{tab:regform}
\vspace{-1mm}
\resizebox{\linewidth}{!}{
\renewcommand{\tabcolsep}{4mm}
\begin{tabular}{lccccc}
\toprule
$\mathcal{L}_{reg}$ form on teacher input $[z_{q_l}, z_{a_l}]$ & MMB & MathVista & MM-Vet & MMMU & Avg \\
\midrule
None                                                                               & $88.2$ & $69.8$ & $63.5$ & $58.9$ & $70.1$ \\
Feature MSE: $\|\textit{Recalibrator}([z_{q_l}, z_{a_l}]) - [z_{q_l}, z_{a_l}]\|^2$ & $88.5$ & $71.2$ & $65.3$ & $61.8$ & $71.7$ \\
KL only via $\textit{VLM-head}_l$                                                   & $88.8$ & $72.6$ & $68.5$ & $64.0$ & $73.5$ \\
CE only via $\textit{VLM-head}_l$                                                   & $89.1$ & $73.7$ & $69.8$ & $65.6$ & $74.6$ \\
\rowcolor{colorful}
\textbf{Ours (\cref{alg:2}): CE $+$ KL via $\textit{VLM-head}_l$}                   & \textbf{89.5} & \textbf{74.9} & \textbf{73.2} & \textbf{68.1} & \textbf{76.4} \\
\bottomrule
\end{tabular}
}
\end{table}


\clearpage
\section{Error Profile and Calibration Analysis}
\phantomsection\label{app:I}
Beyond aggregate benchmark scores, we analyze \emph{how} GenRecal errs relative to the teacher, pooled over MathVista~\cite{lu2023mathvista}, MM-Vet~\cite{yu2023mm}, MMMU~\cite{yue2023mmmu}, and AI2D~\cite{kembhavi2016diagram}. In \cref{tab:errprof}, ``Err.'' is the number of wrong answers; the category columns (Perception/OCR, Spatial, Calculation, Reasoning, Hallucination) report the \% distribution among errors; JSD is the Jensen--Shannon divergence between each model's error distribution and the teacher's; ECE is the expected calibration error; and WConf.\ is the average confidence on wrong answers. GenRecal is markedly closer to the teacher than SFT-only in both \emph{what} it gets wrong (lower JSD) and \emph{how confident} it is when wrong (lower ECE and WConf.), indicating that distillation transfers the teacher's error structure and calibration, not merely its accuracy.

\begin{table}[h!]
\centering
\caption{Error-profile and calibration analysis pooled over MathVista / MM-Vet / MMMU / AI2D. Category columns are \% among wrong predictions. JSD / ECE / WConf.: lower is better; JSD is reported as $10^2\times$ the raw value, and WConf.\ is the average confidence on wrong answers.}
\label{tab:errprof}
\vspace{-1mm}
\resizebox{\linewidth}{!}{
\renewcommand{\tabcolsep}{4mm}
\begin{tabular}{lccccccccc}
\toprule
Method & Err. & Perc/OCR & Spat. & Calc. & Reas. & Hall. & JSD$\times10^2\downarrow$ & ECE$\downarrow$ & WConf.$\downarrow$ \\
\midrule
SFT-only & $520$ & $28$ & $14$ & $22$ & $23$ & $13$ & $0.53$ & $0.21$ & $0.71$ \\
\rowcolor{colorful}
GenRecal & $360$ & $34$ & $16$ & $19$ & $19$ & $12$ & $\mathbf{0.02}$ & $\mathbf{0.13}$ & $\mathbf{0.58}$ \\
\midrule
Teacher  & $340$ & $35$ & $17$ & $18$ & $18$ & $12$ & $0$ & $0.11$ & $0.55$ \\
\bottomrule
\end{tabular}
}
\end{table}


\clearpage
\section{Compute and Baseline-Fairness Comparison}
\phantomsection\label{app:J}
To address baseline fairness, we emphasize that all distillation baselines in \cref{tab:4} use the identical 9M corpus and the same optimization setup within each comparable stage; differences in the number of stages follow the original recipe of each method. \cref{tab:fairness} summarizes per-method compute (per-step model-pass FLOPs) and approximate wall-clock time on the Qwen2-VL-7B student with the Qwen2-VL-72B teacher. Stage~1 of GenRecal is lightweight because only the $\sim$525M Recalibrator (\cref{tab:5}(a)) receives gradients while the rest is a forward pass. As a result, despite its three-stage recipe and extra module, GenRecal's total cost is comparable to---or lower than---the distillation baselines, while achieving substantially higher average performance.

\begin{table}[h!]
\centering
\caption{Compute and baseline-fairness comparison on the Qwen2-VL-7B student with the Qwen2-VL-72B teacher. FLOPs are per-step model-pass FLOPs (same convention as \cref{tab:5}(a)); wall-clock time is approximate. Avg.\ is over MMB / MathVista / MM-Vet / MMMU.}
\label{tab:fairness}
\vspace{-1mm}
\resizebox{\linewidth}{!}{
\renewcommand{\tabcolsep}{5mm}
\begin{tabular}{lllccc}
\toprule
Method & Recipe & Data & Avg & FLOPs ($\times10^{12}$) & Time \\
\midrule
SFT-only                 & 1-stage & $9$M           & $66.3$ & $25.7$  & $\sim$$8$h \\
MiniLLM                  & 2-stage & $9$M/$9$M      & $67.1$ & $311.4$ & $\sim$$21$h \\
DistiLLM                 & 2-stage & $9$M/$9$M      & $67.4$ & $287.0$ & $\sim$$17$h \\
LLaVA-KD                 & 3-stage & $9$M/$9$M/$6$M & $67.7$ & $285.7$ & $\sim$$24$h \\
\rowcolor{colorful}
\textbf{GenRecal (Ours)} & 3-stage & $9$M/$9$M/$6$M & \textbf{72.3} & $288.0$ & $\sim$$18$h \\
\bottomrule
\end{tabular}
}
\end{table}


\end{document}